\documentclass[letterpaper,twocolumn,10pt]{article}
\usepackage{usenix2019} %

\usepackage{amsmath,amssymb,amsfonts}
\usepackage{graphicx}
\usepackage{textcomp}
\usepackage{xspace}
\usepackage{xcolor}
\usepackage{pifont}
\usepackage{textcomp}
\usepackage{relsize}
\usepackage[numbers,square,sort&compress]{natbib}

\usepackage[font=footnotesize]{subcaption}

\usepackage{listings}
\usepackage[font=footnotesize]{caption}

\usepackage[]{algorithm}
\usepackage[noend]{algpseudocode}

\usepackage{amsthm}
\theoremstyle{definition}
\newtheorem{definition}{Definition}[]

\usepackage{booktabs}
\usepackage{tabularx}
\usepackage{multirow}

\usepackage{microtype}

\DeclareMathAlphabet{\mathcal}{OMS}{cmsy}{m}{n}

\renewcommand{\paragraph}[1]{{\vskip 8pt \noindent\bf #1 }}

\newcommand{\head}[1]{\textnormal{\textbf{#1}}}

\newcommand{\code}[1]{{\smaller\texttt{#1}}}
\newcommand{\codeintext}[1]{\mbox{\textproc{\smaller #1}}}

\newcommand{\dingsThree}{\textcolor{sunygreen!50!black}{\ding{183}} }
\newcommand{\dingsTwo}{\textcolor{sunygreen!50!black}{\ding{182}} }

\DeclareMathOperator*{\argmax}{\textrm{arg\,max}}

\definecolor{goeblue}{RGB}{0,51,102}
\definecolor{denceblue}{RGB}{6,107,176}
\definecolor{sunygreen}{RGB}{0,166,77}

\usepackage{tikz}
\usepackage{blkarray} %
\usepackage{pgfplots}
\pgfplotsset{compat=1.11}
\usetikzlibrary{calc}
\usetikzlibrary{positioning,arrows,fit,matrix}
\usetikzlibrary{backgrounds}
\usetikzlibrary{patterns}
\usetikzlibrary{shapes.arrows} %
\usepgfplotslibrary{groupplots}
\usepgfplotslibrary{statistics}

\tikzset{
	>=stealth',
	font=\footnotesize
}

\begin{document}
\date{}

\title{Misleading Authorship Attribution of Source Code\\using
Adversarial Learning\thanks{Published at {USENIX} Security Symposium 
	2019}}

\author{
	{\rm Erwin Quiring, Alwin Maier and Konrad Rieck}\\[3mm]
	\begin{minipage}{8cm}
		\centering \it
		Technische Universit\"at Braunschweig, Germany
	\end{minipage}
}

\maketitle

\subsection*{Abstract}

In this paper, we present a novel attack against authorship
attribution of source code. 
We exploit that recent attribution methods rest on machine learning and
thus can be deceived by adversarial examples of source code.
Our attack performs a series of semantics-preserving 
code transformations that mislead learning-based attribution but appear 
plausible to a developer.
The attack is guided by Monte-Carlo tree search that enables us to 
operate in the discrete domain of source code.
In an empirical evaluation with source code from 204~programmers, we
demonstrate that our attack has a substantial effect on two recent
attribution methods, whose accuracy drops from over 88\% to 1\% under
attack. Furthermore, we show that our attack can imitate the coding
style of developers with high accuracy and thereby induce false
attributions.
We conclude that current approaches for authorship attribution are
inappropriate for practical application and there is a need for
resilient analysis techniques.

\section{Introduction}

The source code of a program often contains peculiarities that reflect
individual coding style and can be used for identifying the
programmer.  These peculiarities---or \emph{stylistic
  patterns}---range from simple artifacts in comments and code layout
to subtle habits in the use of syntax and control flow. A programmer
might, for example, favor while-loops even though the use of for-loops
would be more appropriate.  The task of identifying a programmer based
on these stylistic patterns is denoted as \emph{authorship
  attribution}, and several methods have been proposed to recognize
the authors of source code \citep[][]{AbuAbuMoh+18, CalHarLiuNar+15,
  AlsDauHarMan+17, FraStaGri+06} and compiled programs
\citep[][]{RosZhuMil11, CalYamTau+18, AlrShiWanDeb+18, MenMilJun17}.

While techniques for authorship attribution have made great progress in
the last years, their robustness against attacks has received only
little attention so far, and the majority of work has focused on
achieving high accuracy. The recent study by \citet{SimZetKoh18},
however, shows that developers can manually tamper with the
attribution of source code and thus it becomes necessary to reason
about attacks that can forge stylistic patterns and mislead
attribution methods.

\begin{figure}[bp]
	\vspace{-1.205em}
	\includegraphics[]{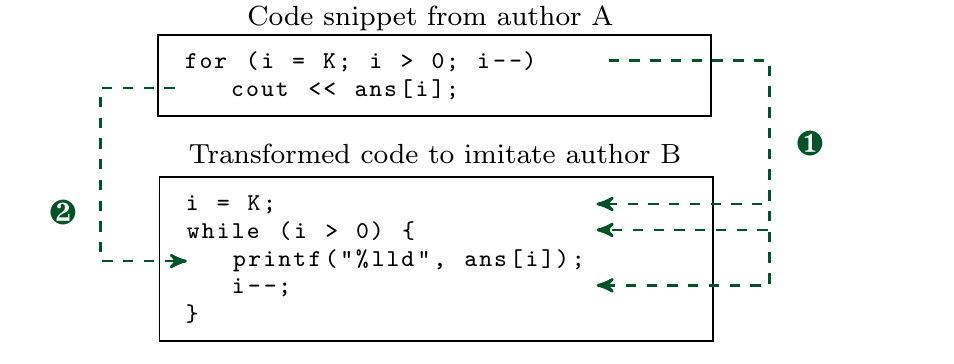}
	\vspace{-1.75em}
	\caption{Two iterations of our attack: Transformation
          \textcolor{sunygreen!50!black}{\ding{182}} changes the control
          statement \code{for} $\rightarrow$ \code{while} and
          transformation \textcolor{sunygreen!50!black}{\ding{183}}
          manipulates the API usage \code{ostream} $\rightarrow$
          \code{printf} to imitate the stylistic patterns of author
          B.}
	\label{fig: intro-coding-example}
\end{figure}

In this paper, we present the first black-box attack against
authorship attribution of source code. Our attack exploits that recent
attribution methods employ machine learning %
and thus can be vulnerable to adversarial
examples~\citep[see][]{PapMcDSinWel+18}.
We combine concepts from adversarial learning and compiler
engineering, and create adversarial examples in the
space of semantically-equivalent programs.

Our attack proceeds by iteratively transforming the source code of a
program, such that stylistic patterns are changed while the underlying
semantics are preserved. To determine these transformations, we
interpret the attack as a game against the attribution method and
develop a variant of \mbox{Monte-Carlo tree
search}~\citep{SilHuaMad+16} for
constructing a sequence of adversarial but plausible
transformations. This black-box strategy enables us to construct
\emph{untargeted attacks} that thwart a correct attribution as well as
\emph{targeted attacks} that imitate the stylistic patterns of a
developer.

As an example, Figure~\ref{fig: intro-coding-example} shows two
transformations performed by our attack on a code snippet from the
Google Code Jam competition. The first transformation changes the for-loop
to a while-loop, while the second replaces the C++ operator \code{<<}
with the C-style function \code{printf}. Note that the format string
is automatically inferred from the variable type.  Both
transformations change the stylistic patterns of author A and, in
combination, mislead the attribution to author B.

We conduct a series of experiments to evaluate the efficacy of our
attack using the source code of 204 programmers from the Google Code
Jam competition. As targets we consider the recent attribution methods
by \citet{CalHarLiuNar+15} and \citet{AbuAbuMoh+18}, which provide
superior performance compared to related approaches.  In our first
experiment, we demonstrate that our attack considerably affects both
attribution methods \citep{AbuAbuMoh+18,CalHarLiuNar+15}, whose
accuracy drops from over 88\% to~1\% under attack, indicating that
authorship attribution can be automatically thwarted at large
scale. In our second experiment, we investigate the effect of targeted
attacks. We show that in a group of programmers, each individual can
be impersonated by 77\% to 81\% of the other developers on
average. Finally, we demonstrate in a study with 15~participants that
code transformed by our attack is plausible and hard to discriminate
from unmodified source code.

Our work has implications on the applicability of authorship
attribution in practice: We find that both, untargeted and targeted
attacks, are effective, rendering the reliable identification of
programmers questionable.  Although our approach builds on a fixed set
of code transformations, we conclude that features regularly
manipulated by compilers, such as specific syntax and control flow,
are not reliable for constructing attribution methods. As a
consequence, we suggest to move away from these features and seek for
more reliable means for identifying authors in source code.

\paragraph{Contributions.}  In summary, we make the following major
contributions in this paper:
\begin{itemize}  \setlength{\itemsep}{-1pt}

\item \emph{Adversarial learning on source code.} We present the first
automatic attack against authorship attribution of source code. We 
consider targeted as well as untargeted attacks of the
attribution method.

\item \emph{Monte-Carlo tree search.}
We introduce Monte-Carlo tree search as a novel approach to guide
the creation of adversarial examples, such that
feasibility constraints in the domain of source code are satisfied.

\item \emph{Black-box attack strategy.} The devised attack does not
require internal knowledge of the attribution method, so that it is
applicable to any learning algorithm and suitable for evading a wide
range of attribution methods.

\item \emph{Large-scale evaluation.} We empirically evaluate our
  attack on a dataset of 204 programmers and demonstrate that
  manipulating the attribution of source code is possible in the
  majority of the considered cases.
\end{itemize}

The remainder of this paper is organized as follows: We review the 
basics
of program authorship attribution in Section \ref{sec:background}. The
design of our attack is lay out in Section \ref{sec:attack}, while
Section~\ref{sec:transformations} and \ref{sec:attackstrategy} discuss
technical details on code transformation and adversarial learning,
respectively. An empirical evaluation of our attack is presented in
Section~\ref{sec:eval} along with a discussion of limitations in
Section \ref{sec:limitations}. Section \ref{sec:related-work}
discusses related work and Section \ref{sec:conclusion} concludes the
paper.

\section{Authorship Attribution of Source Code}
\label{sec:background}

Before introducing our attack, we briefly review the design of
methods for authorship attribution. To this end, we denote the source
code of a program as $x$ and refer to the set of all possible source
codes by $\mathcal{X}$.  Moreover, we define a finite set of authors
$\mathcal{Y}$.  Authorship attribution is then the task of identifying
the author $y \in \mathcal{Y}$ of a given source code
$x \in \mathcal{X}$ using a classification function $f$ such
that~$f(x) = y$. In line with the majority of previous work, we assume
that the programs in $\mathcal{X}$ can be attributed to a single
author, as the identification of multiple authors is an ongoing
research effort \citep[see][]{MenMilJun17, DauCalHarGre17}.

Equipped with this basic notation, we proceed to discuss the two main
building
blocks of current methods for authorship attribution: (a) the
extraction of features from source code and (b) the application of
machine learning for constructing the classification function.

\subsection{Feature Extraction}

The coding habits of a programmer can manifest in a variety of
stylistic patterns. Consequently, methods for authorship attribution
need to extract an expressive set of features from source code that
serve as basis for inferring these patterns.  In the following, we
discuss the major types of these features and use the code sample in
Figure \ref{code:background-code-example} as a running example
throughout the paper.

\begin{figure}[htbp]
	\begin{lstlisting}[
	xleftmargin=.14\columnwidth, %
	xrightmargin=.14\columnwidth, %
	numbers=left,
	framexleftmargin=2pt,
	]
int foo(int a){
	int b;
	if (a < 2)      // base case
		return 1;
	b = foo(a - 1); // recursion
	return a * b;
}
	\end{lstlisting}
	\vspace{-0.75em}
	\caption{Exemplary code sample (see Figure \ref{fig:background-ast},
		\ref{fig:cfg-udc}, and  \ref{fig:ast-drm})}
	\label{code:background-code-example}
\end{figure}

\paragraph{Layout features.} Individual preferences of a programmer
often manifest in the layout of the code and thus corresponding
features are a simple tool for characterizing coding style. Examples
for such features are the indentation, the form of comments and the
use of brackets. In Figure~\ref{code:background-code-example}, for
instance, the indentation width is 2, comments are provided in C++
style, and curly braces are opened on the same line.

Layout features are trivial to forge, as they can be easily modified
using tools for code formatting, such as GNU indent. Moreover, many
integrated development editors automatically normalize source code,
such that stylistic patterns in the layout are unified. 

\paragraph{Lexical features.} A more advanced type of features can be
derived from the lexical analysis of source code. In this analysis
stage, the source code is partitioned into so-called \emph{lexems},
tokens that are matched against the terminal symbols of the language
grammar. These lexems give rise to a strong set of string-based
features jointly covering keywords and symbols. For example, in
Figure~\ref{code:background-code-example}, the frequency of the lexem
\code{int} is~3, while it is 2 for the lexem \code{foo}.

In contrast to code layout, lexical features cannot be easily
manipulated, as they implicitly describe the syntax and semantics of
the source code.  While the lexem \code{foo} in the running example
could be easily replaced by another string, adapting the lexem
\code{int} requires a more involved code transformation that
introduces a semantically equivalent data type. We introduce such a
transformation in Section~\ref{sec:transformations}.

\begin{figure}[tbp]
	\centering
	\includegraphics[width=0.48\textwidth]{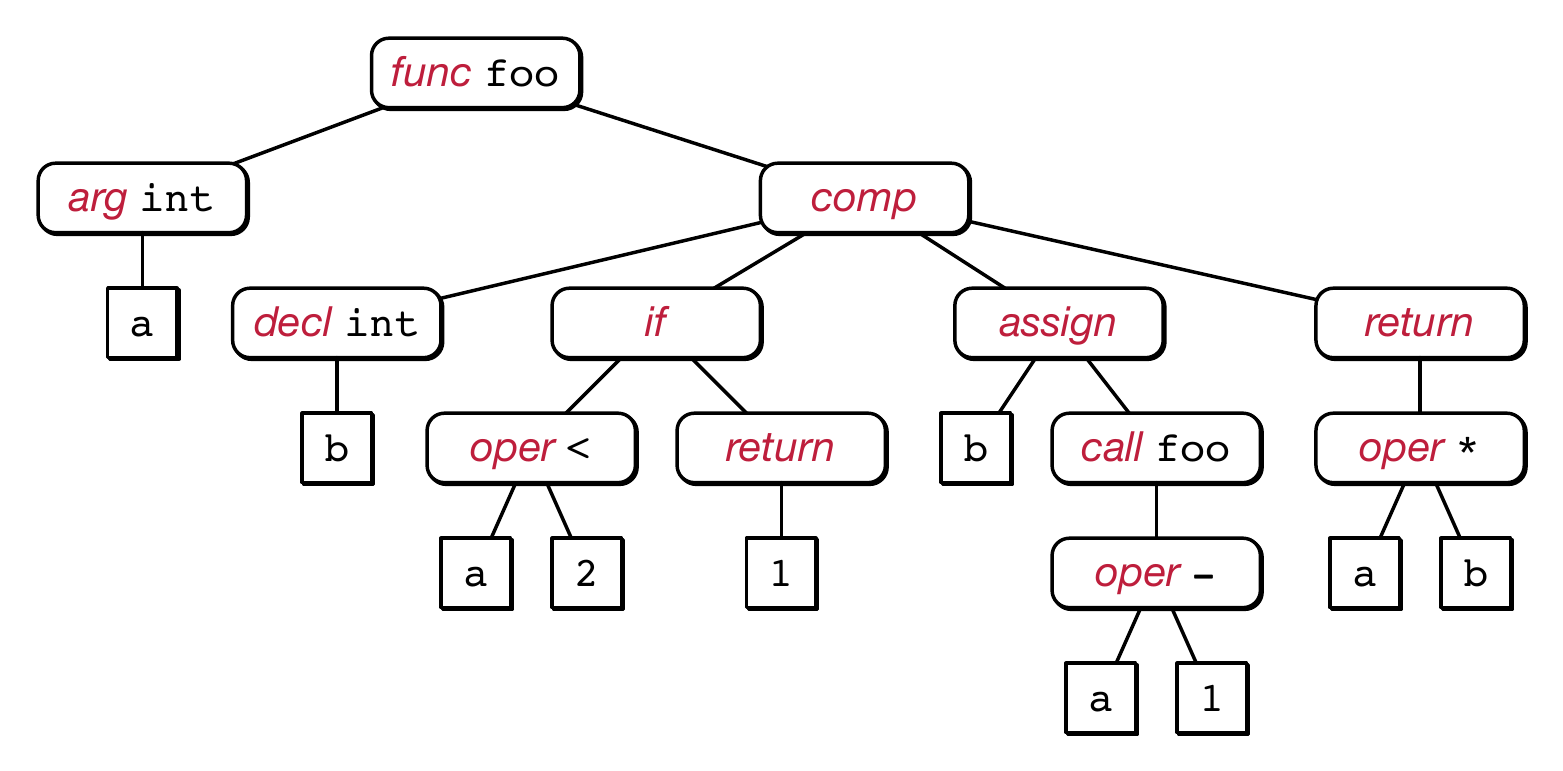}
	\vspace{-1.995em}
	\caption{Abstract syntax tree (AST) for code sample in
		Figure~\ref{code:background-code-example}. }
	\label{fig:background-ast}
\end{figure}

\paragraph{Syntactic features.} The use of syntax and control flow
also reveals individual stylistic patterns of programmers.  These
patterns are typically accessed using the \emph{abstract syntax tree}
(AST), a basic data structure of compiler design \cite{AhoSetUll06}.
As an example, Figure~\ref{fig:background-ast} shows a simplified AST
of the code snippet from Figure~\ref{code:background-code-example}.
The AST provides the basis for constructing an extensive set of
syntactic features. These features can range from the specific use of
syntactic constructs, such as unary and ternary operators, to generic
features characterizing the tree structure, such as the frequency of
adjacent nodes.  In Figure~\ref{fig:background-ast}, there exist 21
pairs of adjacent nodes including, for example, \code{(func
  foo)}$\rightarrow$\code{(arg int)} and
\code{(return)}$\rightarrow$\code{(1)}.

Manipulating features derived from an AST is challenging, as even
minor tweaks in the tree structure can fundamentally change the
program semantics. As a consequence, transformations to the AST need
to be carefully designed to preserve the original semantics and to avoid
unintentional side effects. For example, removing the node pair
\code{(decl int)}$\rightarrow$\code{(b)} from the AST in
Figure~\ref{fig:background-ast} requires either replacing the type or
the name of the variable without interfering with the remaining
code. In practice, such transformations are often non-trivial and we
discuss the details of manipulating the AST in
Section~\ref{sec:transformations}.

\subsection{Machine Learning}

The three feature types (layout, lexical, syntactic) provide a broad
view on the characteristics of source code and are used by many
attribution methods as the basis for applying machine-learning
techniques \citep[e.g.,][]{CalHarLiuNar+15, AbuAbuMoh+18,
  AlsDauHarMan+17, Pel00}

\paragraph{From code to vectors.} Most learning algorithms are
designed to operate on vectorial data and hence the first step for
application of machine learning is the mapping of code to a vector
space using the extracted features. Formally, this mapping can be
expressed as
$
\phi: \mathcal{X} \longrightarrow \mathcal{F} = \mathbb{R}^d
$
where $\mathcal{F}$ is a $d$~dimensional vector space describing
properties of the extracted features. Different techniques can be
applied for constructing this map, which may include the computation
of specific metrics as well as generic embeddings of features and
their relations, such as a TF-IDF weighting \cite{CalHarLiuNar+15,
  AbuAbuMoh+18}.

Surprisingly, the feature map $\phi$ introduces a non-trivial hurdle
for the construction of attacks. The map $\phi$ is usually not
bijective, that is, we can map a given source code~$x$ to a feature
space but are unable to automatically construct the source code $x'$
for a given point $\phi(x')$. Similarly, it is difficult to predict
how a code transformation $x \mapsto x'$ changes the position in
feature space $\phi(x) \mapsto \phi(x')$. We refer to this problem
as the \emph{problem-feature space dilemma} and discuss its
implications in Section~\ref{sec:attack}.

\paragraph{Multiclass classification.}
Using a feature map $\phi$, we can apply machine learning for
identifying the author of a source code. Typically, this is done by
training a \emph{multiclass classifier}
$g: \mathcal{X} \longrightarrow \mathbb{R}^{|\mathcal{Y}|}$
that returns scores for all authors~$\mathcal{Y}$. An
attribution is obtained by simply computing
\[
f(x) = \argmax_{y \in \mathcal{Y}} g_y(x) .
\]
This setting has different advantages: First, one can investigate all
top-ranked authors. Second, one can interpret the returned scores for
determining the confidence of an attribution. We make use of the latter
property for guiding our attack strategy and generating adversarial
examples of source code (see Section \ref{sec:attackstrategy})

Different learning algorithms have been used for constructing the
multiclass classifier $g$, as for example, support vector
machines~\citep{Pel00}, random forests~\citep{CalHarLiuNar+15}, and
recurrent neural networks~\citep{AlsDauHarMan+17,
  AbuAbuMoh+18}. Attacking each of these learning algorithms
individually is a tedious task and thus we resort to a \emph{black-box
  attack} for misleading authorship attribution. This attack does not
require any knowledge of the employed learning algorithm and operates
with the output $g(x)$ only. Consequently, our approach is agnostic to
the learning algorithm as we demonstrate in the evaluation in
Section~\ref{sec:eval}.

\section{Misleading Authorship Attribution}
\label{sec:attack}

With a basic understanding of authorship attribution, we are ready to
investigate the robustness of attribution methods and to develop a
corresponding black-box attack. To this end, we first define our threat
model and attack scenario before discussing technical details in the
following sections.

\subsection{Threat Model}
\label{sec:attackobjectives}

For our attack, we assume an adversary who has black-box access to an
attribution method. That is, she can send an arbitrary source code $x$
to the method and retrieve the corresponding prediction $f(x)$ along
with prediction scores $g(x)$. The training data, the extracted
features, and the employed learning algorithm, however, are unknown to
the adversary, and hence the attack can only be guided by iteratively
probing the attribution method and analyzing the returned prediction
scores. This setting resembles a classic \emph{black-box attack} as
studied by \citet{TraZhaJuel+16} and \citet{PapMcDGoo+16}.
As part of our threat model, we consider two types of
attacks---\emph{untargeted} and \emph{targeted attacks}---that require
different capabilities of the adversary and have distinct implications
for the involved programmers.

\paragraph{Untargeted attacks.} In this setting, the adversary tries to
mislead the attribution of source code by changing the classification
into \emph{any} other programmer.  This attack is also denoted as
\emph{dodging} \citep{ShaBhaBauRei16} and impacts the correctness of
the attribution. As an example, a benign programmer might use this
attack strategy for concealing her identity before publishing the
source code of a program.

\paragraph{Targeted attacks.} The adversary tries to change the
classification into a chosen \emph{target} programmer. This attack
resembles an \emph{impersonation} and is technically more advanced, as
we need to transfer the stylistic patterns from one developer to
another. A targeted attack has more severe implications: A malware
developer, for instance, could systematically change her source code
to blame a benign developer.

Furthermore, we consider two scenarios for targeted attacks: In the
first scenario, the adversary has no access to source code from the
target programmer and thus certain features, such as variable names
and custom types, can only be guessed. In the second
scenario, we assume that the adversary has access to two files of
source code from the target developer. Both files are not part of the
training- or test set and act as external source for extracting template
information, such as recurring custom variable names.

\vspace{6pt}
\noindent
In addition, we test a scenario where the targeted attack solely rests 
on a separate training set, 
without access to the output of the original classifier.
This might be the case, for instance, if the attribution method is 
secretly deployed, but code samples are available from public code 
repositories.
In this scenario, the adversary can learn a substitute model with the 
aim that her adversarial example---calculated on the 
substitute---also transfers to the original classifier.

\subsection{Attack Constraints}
\label{sec:attackconstraints}

Misleading the attribution of an author can be achieved with different
levels of sophistication. For example, an adversary may simply copy
code snippets from one developer for impersonation or heavily
obfuscate source code for dodging. These trivial attacks, however,
generate implausible code and are easy to detect. As a consequence, we
define a set of constraints for our attack that should make it hard to
identify manipulated source code.

\paragraph{Preserved semantics.}
We require that source code generated by our attack is semantically
equivalent to the original code. That is, the two codes produce
identical outputs given the same input. As it is undecidable whether
two programs are semantically equivalent, we take care of this
constraint during the design of our code transformations and ensure
that each transformation is as semantics-preserving as possible.

\paragraph{Plausible code.}
We require that all transformations change the source code, such that
the result is syntactically correct, readable and plausible. 
The latter constraint corresponds to the aspect of imperceptibility 
when adversarial examples are generated in the image 
domain~\citep{CarWag17}. In our context, plausibility is important 
whenever the adversary wants to hide the modification of a source file, 
for instance, when blaming another developer.
For this reason, we do not include junk code or unusual syntax that 
normal developers would not use.

\paragraph{No layout changes.}
Layout features such as the tendency to start lines with spaces or
tabs are trivial to change with tools for code
formatting (see Section~\ref{sec:pers-semant-plaus}). Therefore, we
restrict our attack to the forgery of lexical and
syntactic features of source code. In this way, we examine our approach
under a more difficult scenario for the attacker where no layout
features are exploitable to mislead the attribution.

\subsection{Problem-Feature Space Dilemma}
\label{sec:challenges}

The described threat model and attack constraints pose unique
challenges to the design of our attack. Our attack jointly operates in
two domains: On the one hand, we aim at attacking a classifier in the
feature space $\mathcal{F}$. On the other hand, we require the source
code to be semantically equivalent and plausible in the problem space
$\mathcal{X}$. For most feature maps~$\phi$, a one-to-one
correspondence, however, does not exist between the two spaces and
thus we encounter a~dilemma.

\paragraph{Problem space $\rightsquigarrow$ feature space.} Each
change in the source code $x$ may impact a set of features in
$\phi(x)$. The exact amount of change is generally not
controllable. The correlation of features and post-processing steps in
$\phi$, such as a TF-IDF weighting, may alter several features, even
if only a single statement is changed in the source code.  This
renders target-oriented modification of the source code difficult.

For example, if the declaration of the variable \code{b} in line 2 of
Figure~\ref{code:background-code-example} is moved to line 5, a series
of lexical and syntactic features change, such as the frequency of the
lexem \code{b} or the subtree under the node \code{assign} in
Figure~\ref{fig:background-ast}.

\paragraph{Feature space $\rightsquigarrow$ problem space.}  Any
change to a feature vector $\phi(x)$ must ensure that there exists a
plausible source code $x$ in the problem space. Unfortunately,
determining $x$ from $\phi(x)$ is not tractable for non-bijective
feature maps, and it is impossible to directly apply techniques
from adversarial learning that operate in the feature space.

For example, if we calculate the difference of two vectors
\mbox{$\phi(z) = \phi(x) - \phi(x') $}, we have no means for
determining the resulting source code $z$.  Even worse, it might be
impossible to construct $z$, as the features in $\phi(z)$ can violate
the underlying programming language specification, for example,
due to feature combinations inducing impossible AST~edges.

This dilemma has received little attention in the literature on
adversarial learning so far, and it is often assumed that an adversary
can change features almost arbitrarily~\citep[e.g.][]{BigCorMai+13,
  PapMcDJhaFreCelSwa16, CarWag17}. Consequently, our attack does not
only pinpoint weaknesses in authorship attribution but also illustrates 
how adversarial learning can be conducted when the problem and feature
space are disconnected.

\subsection{Our Attack Strategy}
\label{sec:ourapproach}

To tackle this challenge, we adopt a mixed attack strategy that
combines concepts from compiler engineering and adversarial
learning. For the problem space, we develop code transformations
(source-to-source compilations) that enable us to maneuver in the
problem space and alter stylistic patterns without changing the
semantics.
For the feature space, we devise a variant of Monte-Carlo
tree search that guides the transformations towards a target. This
variant considers the attack as a game against the attribution method
and aims at reaching a desired output with few transformations.

\begin{figure}[tbp]
	\centering\vspace{-6pt}
	\includegraphics[width=0.49\textwidth]{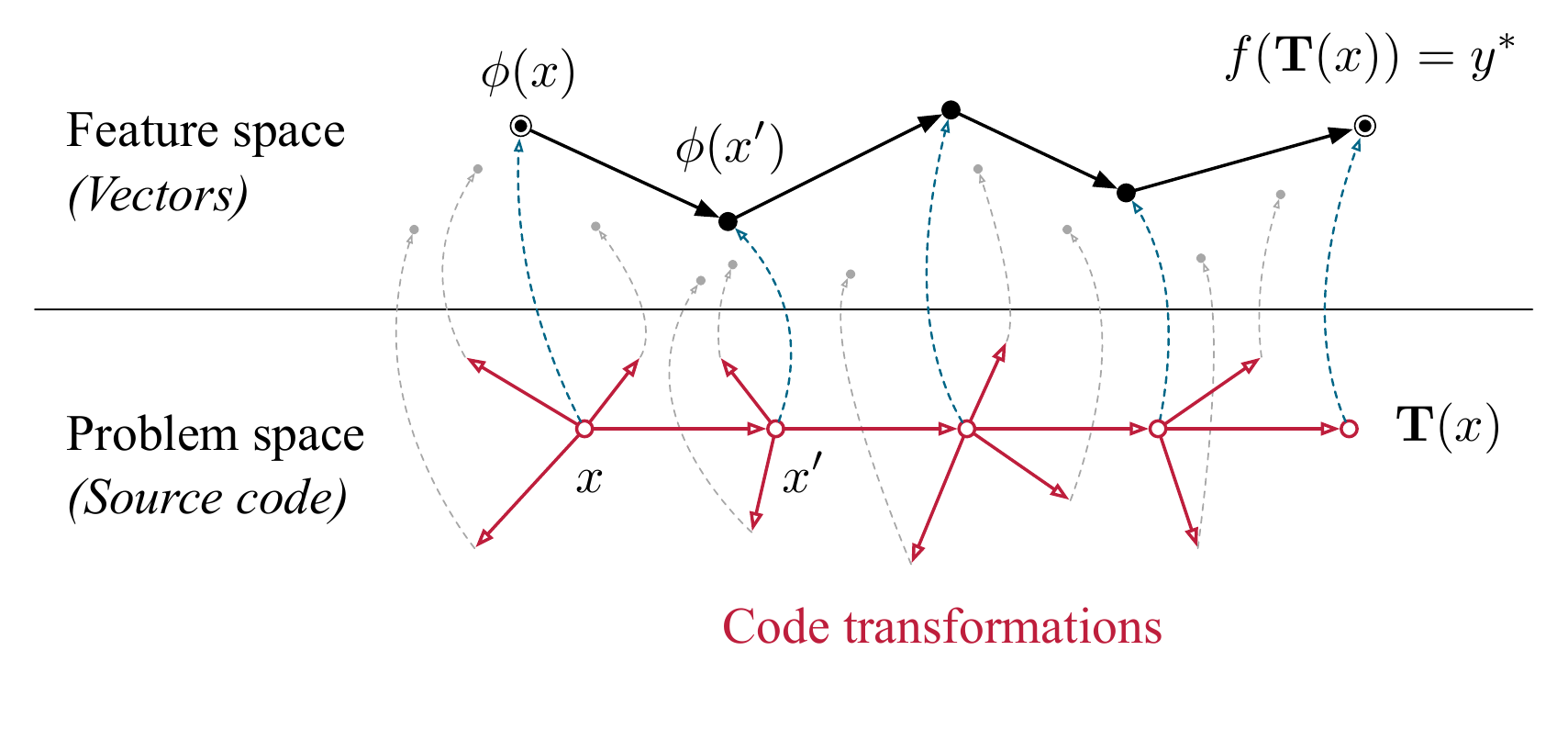}
	\vspace{-25pt}
	\caption{Schematic depiction of our approach. The attack is
          realized by moving in the problem space using code
          transformations while being guided by Monte-Carlo tree
          search in the feature space.}
	\label{fig:overview}
\end{figure}

An overview of our attack strategy is illustrated in
Figure~\ref{fig:overview}. As the building blocks of our approach
originate from different areas of computer science, we discuss their
technical details in separate sections.
First, we introduce the concept of semantics-preserving code
transformations and present five families of source-to-source
transformations~(Section~\ref{sec:transformations}).  Then, we
introduce Monte-Carlo tree search as a generic black-box attack for
chaining transformations together such that a target in the feature
space is reached~(Section~\ref{sec:attackstrategy}).

\section{Code Transformations}\label{sec:transformations}

The automatic modification of code is a well-studied problem in
compiler engineering and source-to-source
compilation~\citep{AhoSetUll06}.  Consequently, we build our code
transformations on top of the compiler frontend \emph{Clang}
\citep{clang18}, which provides all necessary primitives for parsing,
transforming and synthesizing C/C++ source code.
Note that we do \emph{not} use code obfuscation methods, since their 
changes are (a) clearly visible, and (b) cannot mislead a classifier to 
a targeted author.
Before presenting five families of transformations, we formally define
the task of \emph{code transformation} and introduce additional
program representations.

\begin{definition}
  A code transformation $T : \mathcal{X} \longrightarrow 
  \mathcal{X}$,$\;$ $x \mapsto x'$ takes a source code $x$ and 
  generates a transformed version $x'$, such that $x$ and $x'$ are 
  semantically equivalent.
\end{definition}

While code transformations can serve various purposes in general
\citep[][]{AhoSetUll06}, we focus on \emph{targeted} transformations
that modify only minimal aspects of source code. 
If multiple source locations are applicable for a transformation, we 
use a pseudo-random seed to select
one location. To chain together targeted transformations, we define
\emph{transformation sequences} as follows:

\begin{definition}
	A transformation sequence
	$\mathbf{T} = T_1 \circ T_2 \circ \dots \circ T_n$ is the subsequent
	application of multiple code transformations to a source code $x$.
\end{definition}

To efficiently perform transformations, we make use of different
program representations, where the AST is the most important one.
To ease the realization of involved
transformations, however, we employ two additional program
representations that augment our view on the source code.

\begin{figure}[]
	\centering
	\includegraphics[width=0.42\textwidth]{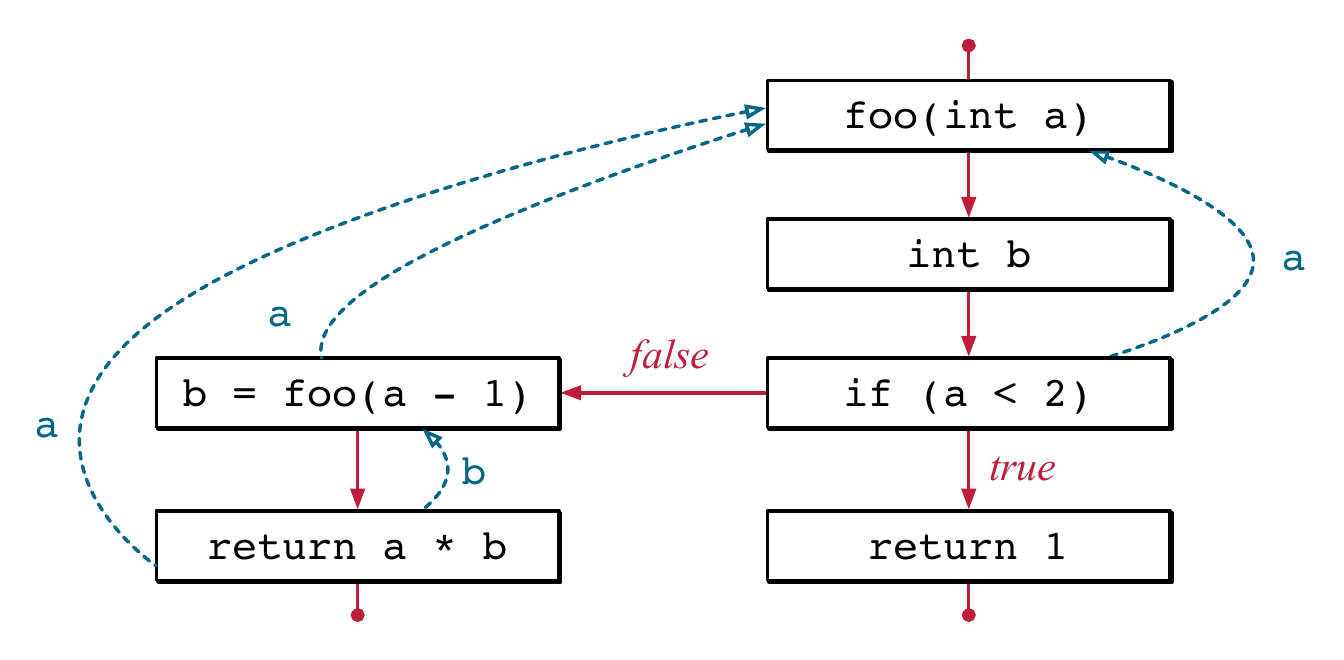}
	\vspace{-5pt}
	\caption{Control-flow graph with use-define chains for
		the code snippet from
		Figure~\ref{code:background-code-example}. The
		control flow is shown in red (solid), use-define
		chains in blue (dashed).}
	\label{fig:cfg-udc}
\end{figure}

\paragraph{Control-flow graph with use-define chains.} The control
flow of a program is typically represented by a \emph{control-flow
  graph} (CFG) where nodes represent statements and edges the flow of
control. Using the CFG, it is convenient to analyze the execution
order of statements.  We further extend the CFG provided by Clang with
\emph{use-define chains} (UDCs).  These chains unveil dependencies
between usages and the definitions of a variable. With the aid of
UDCs, we can trace the flow of data through the program and identify
data dependencies between local variables and function
arguments. Figure~\ref{fig:cfg-udc} shows a CFG with use-define
chains.

\begin{figure}[t]
	\centering
	\includegraphics[width=0.47\textwidth]{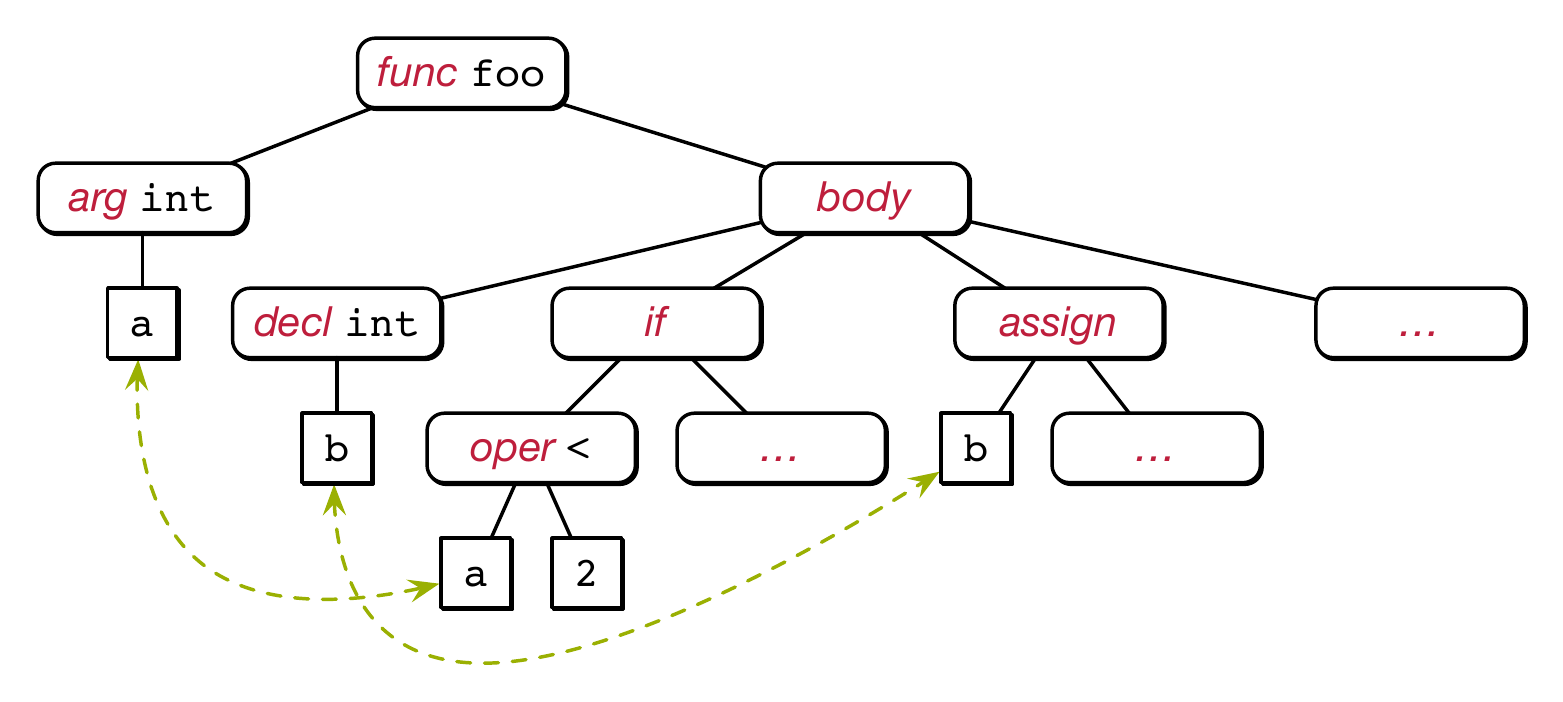}
	\vspace{-7.2pt}
	\caption{Abstract syntax tree with declaration-reference
		mapping for the code snippet from
		Figure~\ref{code:background-code-example}. Declaration
		references are shown in green (dashed).}
	\label{fig:ast-drm}
\end{figure}

\paragraph{Declaration-reference mapping.}
We additionally introduce a declaration-reference mapping (DRM) that
extends the AST and links each declaration to all usages of the
declared variable. As an example, Figure~\ref{fig:ast-drm} shows a
part of the AST together with the respective DRM for the code sample
from Figure~\ref{code:background-code-example}. This code
representation enables navigation between declarations and variables,
which allows us to efficiently rename variables or check for the sound
transformation of data types. Note the difference between
\mbox{use-define} chains and declaration-reference mappings. The
former connects variable usages to variable definitions, while the
latter links variable usages to variable declarations.

Based on these program representations, we develop a set of generic code
transformations that are suitable for changing different stylistic
patterns. In particular, we implement 36~transformers that are
organized into five
families. Table~\ref{tab:code-transformations-overview} provides an
overview of each family together with the program representation used
by the contained transformers.

\begin{table}[tbp]
	\footnotesize
	\centering%
	\renewcommand{\arraystretch}{1.1}%
	\caption{Implemented families of transformations.}
	\label{tab:code-transformations-overview}
	\begin{tabularx}{1\linewidth}{l>{\hsize=0.44\hsize}Xcccc}
		\toprule
		\head{Transformation family} & \head{\#}  & \head{AST} &
		\head{CFG} & \head{UDC} & \head{DRM} \\
		\midrule
		Control transformations     & 5  & $\bullet$ & $\bullet$ &
		$\bullet$ & \\
		Declaration transformations & 14 & $\bullet$ & & & $\bullet$ \\
		API transformations         & 9 & $\bullet$ & $\bullet$ &  &
		$\bullet$ \\
		Template transformations   & 4  & $\bullet$ & & & $\bullet$ \\
		Miscellaneous transformations   & 4  & $\bullet$ & & & \\
		\bottomrule
	\end{tabularx}
\end{table}

In the following, we briefly introduce each of the five families. For
a detailed listing of all 36~transformations, we refer the reader to
Table~\ref{tab:appendix-transf} in Appendix~\ref{sec:appendixA}.

\begin{figure}[h]
	\centering 
	\includegraphics[]{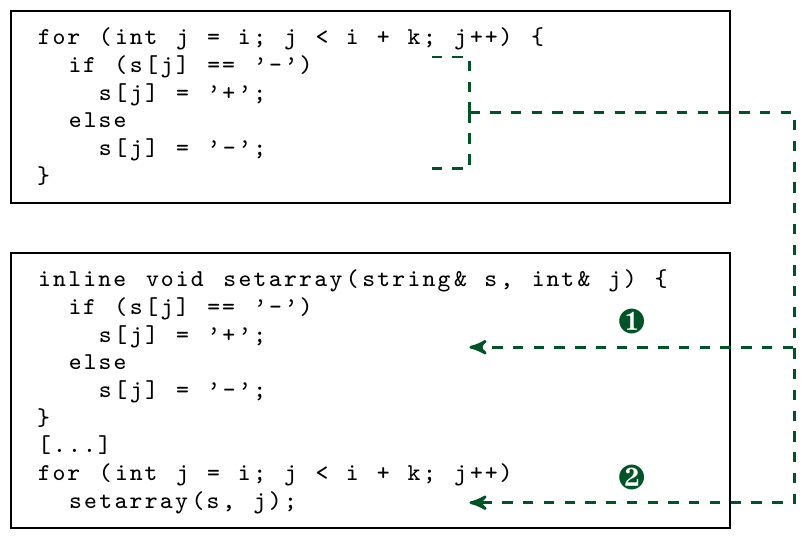}
	\vspace{-0.3em}
	\caption{Example of a control transformation. \dingsTwo moves
          the compound statement into an own function and passes all
          variables defined outside the block as
          parameters. \dingsThree calls the new function at the
          previous location.}
	\label{fig:fctcreation-example}
\end{figure}

\paragraph{Control transformations.}
The first family of source-to-source transformations rewrites
control-flow statements or modifies the control flow between functions.
In total, the family contains 5~transformations.
For example, the control-flow statements \code{while} and \code{for}
can be mutually interchanged by two transformers. These
transformations address a developer's preference to use a particular
iteration type.
As another example, Figure~\ref{fig:fctcreation-example} shows the
automatic creation of a function. The transformer moves the inner
block of the for-statement to a newly created function. This
transformation involves passing variables as function arguments,
updating their values and changing the control flow of the caller and
callee.

\begin{figure}[]
	\centering
\includegraphics[]{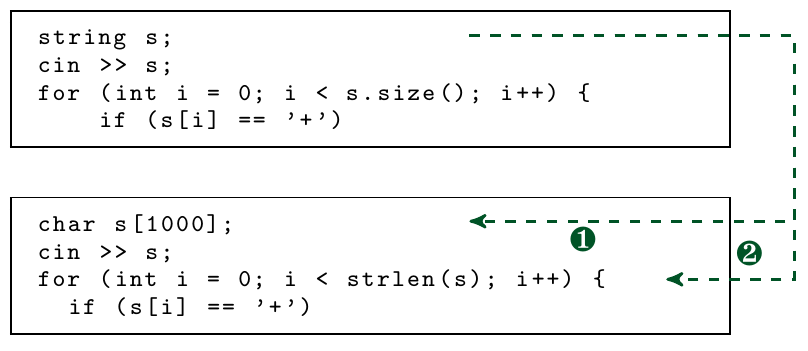}
	\vspace{-0.3em}
	\caption{Example of a declaration transformation. \dingsTwo
		replaces the declaration of the C++ \code{string} object with a
		\texttt{char} array, \dingsThree adapts all uses of the object.}
	\label{fig:datastr-example}
\end{figure}

\paragraph{Declaration transformations.}
This family consists of 14~transformers that modify, add or remove
declarations in source code.
For example, in a widening conversion, the type of a variable is
changed to a larger type, for example, \code{int} to \code{long}. This
rewriting mimics a programmer's preference for particular data
types. Declaration transformations make it necessary to update all
usages of variables which can be elegantly carried out using the DRM
representation.
Replacing an entire data type is a more challenging transformation, as
we need to adapt all usages to the type, including variables,
functions and return values.  Figure~\ref{fig:datastr-example} shows
the replacement of the C++ \code{string} object with a conventional
\code{char} array, where the declaration and also API functions, such
as \code{size}, are modified.  Note that in our current
implementation of the transformer the \code{char} array has a fixed
size and thus is not strictly equivalent to the C++ \code{string}
object.

\paragraph{API transformations.}
The third family contains 9~transformations and exploits the fact
that various APIs can be used to solve the same problem. Programmers
are known to favor different APIs and thus tampering with API usage is
an effective strategy for changing stylistic patterns. For instance,
we can choose between various ways to output information in C++, such
as \code{printf}, \code{cout}, or \code{ofstream}.

\begin{figure}[tbp]
	\centering
\includegraphics[]{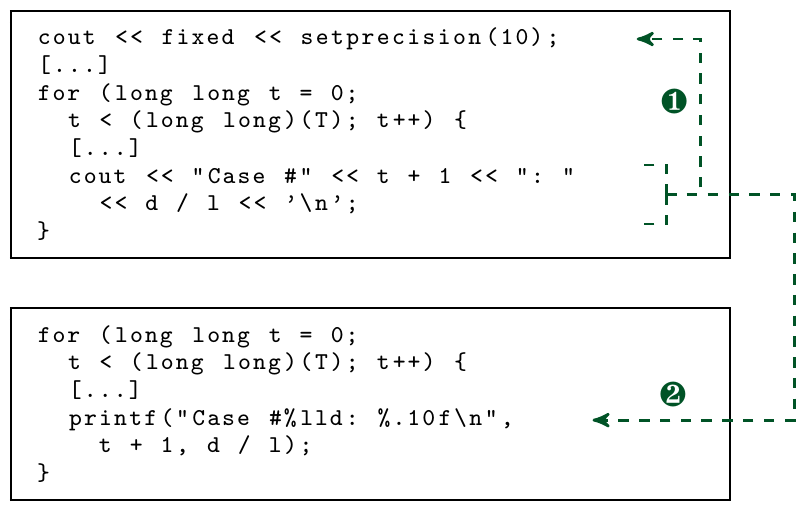}
	\vspace{-0.4em}
	\caption{Example of an API transformation. \dingsTwo
		determines the current precision for output; \dingsThree
		replaces the C++ API with a C-style printf. The format
		specifier respects the precision and the data type of the
		variable.}
	\label{fig:API-example}
\end{figure}

As an example, Figure~\ref{fig:API-example} depicts the replacement of
the object \code{cout} by a call to \code{printf}. To this end, the
transformer first checks for the decimal precision of floating-point
values that \code{cout} employs, that is, we use the CFG to find the
last executed \code{fixed} and \code{setprecision} statement. Next,
the transformer uses the AST to resolve the final data type of each
\code{cout} entry and creates a respective format string for
\code{printf}.

\paragraph{Template transformations.}
The fourth family contains 4~transformations that insert or change
code patterns based on a give template. For example, authors tend to
reuse specific variable names, constants, and type definitions. If a
template file is given for a target developer, these information are
extracted and used for transformations. Otherwise, default values that
represent general style patterns are employed. For instance, variable
names can be iteratively renamed into default names like \code{i},
\code{j}, or \code{k} until a developer's tendency to declare control
statement variables is lost (dodging attack) or gets matched
(impersonation attack).

\paragraph{Miscellaneous transformations.}
The last family covers 4~transformations that conduct generic changes
of code statements. For example, the use of curly braces around
compound statements is a naive but effective stylistic pattern for
identifying programmers.  The compound statement transformer thus
checks if the body of a control statement can be enclosed by curly
braces or the other way round. In this way, we can add or remove a
compound statement in the AST.

Another rather simple stylistic pattern is the use of return
statements, where some programmers omit these statements in the
\code{main} function and others differ in whether they return a
constant, integer or variable. Consequently, we design a transformer
that manipulates return statements.

\section{Monte-Carlo Tree Search}\label{sec:attackstrategy}

Equipped with different code transformations for changing stylistic
patterns, we are ready to determine a sequence of these
transformations for untargeted and targeted attacks.
We aim at a short sequence, which makes the attack less likely to be 
detected.
Formally, our objective is to find a short transformation
sequence~$\mathbf{T}$ that manipulates a source file $x$, such that
the classifier $f$ predicts the target label $y^{*}$:
\begin{equation}
\begin{aligned}
\textstyle f\bigl( \mathbf{T}(x) \bigr) = y^{*} \; .
\end{aligned}
\label{eq:evasion_problem_main}
\end{equation}
In the case of an untargeted attack, $y^{*}$ represents any
other developer than the original author $y^s$, that is,
$y^{*} \neq y^s$.  In the case of a targeted attack, $y^{*}$ is defined 
as a particular target author~$y^t$.

As we are unable to control how a transformation $T(x)$ moves the
feature vector $\phi(x)$, several standard techniques for solving the
problem in \eqref{eq:evasion_problem_main} are not
applicable, such as gradient-based methods~\citep[e.g.][]{CarWag17}.
Therefore, we require an algorithm that works over a search space of
discrete objects such as the different transformations of the source
code. As a single transformation does not necessarily change the score
of the classifier, simple approximation techniques like Hill Climbing
that only evaluate the neighborhood of a sample fail to provide
appropriate solutions.

\begin{figure}
	\centering
\includegraphics[]{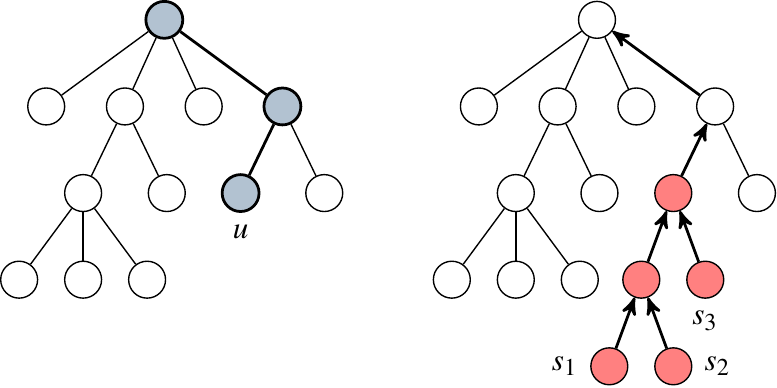}
	\caption{Basic steps of Monte-Carlo tree search.
		The left plot shows the selection step, the right plot the
		simulation, expansion and backpropagation.
	}\label{fig:mcts_figure}
\end{figure}

As a remedy, we construct our attack algorithm around the concept of
\emph{\mbox{Monte-Carlo tree search}} (MCTS)---a strong search
algorithm that has proven effective in AI gaming with AlphaGo
\cite{SilHuaMad+16}. Similar to a game tree, our variant of MCTS
creates a search tree for the attack, where each node represents a
state of the source code and the edges correspond to
transformations. By moving in this tree, we can evaluate the impact of
different transformation sequences before deciding on the next move.
Figure~\ref{fig:mcts_figure} depicts the four basic steps of our 
algorithm: selection, simulation, expansion and backpropagation.

\paragraph{Selection.}
As the number of possible paths in the search tree grows
exponentially, we require a \emph{selection policy} to identify the
next node for expansion.  This policy balances the tree's exploration
and exploitation by alternately selecting nodes that have not been
evaluated much (exploration) and nodes that seem promising to obtain a
better result (exploitation). %
Following this policy, we start at the root node and recursively
select a child node until we find a node~$u$ which was not
evaluated before. %
Appendix~\ref{sec:appendixE} gives more information about the used 
selection policy.

\paragraph{Simulation \& Expansion.}
We continue by generating a set of unique transformation sequences
with varying length that start at~$u$.  We bound the length of each
sequence by a predefined value. In our experiments, we create
sequences with up to 5~transformations. For each sequence, we
determine the classifier score by providing the modified source
code to the attribution method.  The right plot in
Figure~\ref{fig:mcts_figure} exemplifies the step: we create three
sequences where two have the same first transformation. Next, we
create the respective tree nodes. As two sequences start with the same
transformation, they also share a node in the search tree.

\paragraph{Backpropagation.}
As the last step, we propagate the obtained classifier scores from the
leaf node of each sequence back to the root. During this propagation,
we update two statistics in each node on the path: First, we increment
a counter that keeps track of how often a node has been part of a
transformation sequence. In Figure~\ref{fig:mcts_figure}, we increase 
the visit count of node $u$ and the nodes above by~3.
Second, we store the classifier scores in
each node that have been observed in its subtree.  For example,
node~$u$ in Figure \ref{fig:mcts_figure} stores the scores from $s_1$,
$s_2$ and $s_3$. Both statistics are used by the selection policy and 
enable us to balance the exploration and exploitation of the tree in 
the next iterations.

\paragraph{Iteration.}
We repeat these four basic steps until a predefined iteration
constraint is reached.
After obtaining the resulting search tree, we identify the
root's child node with the highest average classifier score and make
it the novel root node of the tree. We then repeat the entire process
again. The attack is stopped
if we succeed, we reach a previously fixed number of iterations, or we 
do not obtain any improvement over multiple iterations.

\smallskip \noindent Appendix~\ref{sec:appendixE} provides more 
implementation details on our variant of MCTS. We finally note that the 
algorithm resembles a black-box attack, as the inner working of the 
classifier $f$ is not considered.

\section{Evaluation}\label{sec:eval}

We proceed with an empirical evaluation of our attacks and investigate 
the robustness of source-code authorship attribution in a series of 
experiments.  In particular, we investigate the impact of untargeted 
and targeted attacks on two recent attribution methods 
(Section~\ref{sec:untargeted-evasion} \& \ref{sec:targeted-evasion}).
Finally, we verify in Section~\ref{sec:pers-semant-plaus} that our
initially imposed attack constraints are fulfilled.

\subsection{Experimental Setup}

Our empirical evaluation builds on the methods developed by
Caliskan et al.~\cite{CalHarLiuNar+15} and \citet{AbuAbuMoh+18}, two 
recent
approaches that operate on a diverse set of features and provide
superior performance in comparison to other attribution methods.  For
our evaluation, we follow the same experimental setup as the authors,
re-implement their methods and make use of a similar dataset.

\paragraph{Dataset \& Setup.}  We collect C++ files from the 2017
\textit{Google Code Jam} (GCJ) programming
competition~\citep{website:googlecodejam}. This contest consists of
various rounds where several participants solve the same programming
challenges. This setting enables us to learn a classifier for
attribution that separates stylistic patterns rather than artifacts of
the different challenges. Moreover, for each challenge, a test input is
available that we can use for checking the program semantics.  Similar
to previous work, we select eight challenges from the competition and
collect the corresponding source codes from all authors who solved
these challenges.

In contrast to prior work~\cite{CalHarLiuNar+15, AbuAbuMoh+18},
however, we set more stringent restrictions on the source code. We
filter out files that contain incomplete or broken
solutions. Furthermore, we format each source code using
\code{clang-format} and expand macros, which removes artifacts that
some authors introduce to write code more quickly during the contest.
Our final dataset consists of 1,632~files of C++ code from 204~authors
solving the same 8~programming challenges of the competition.

For the evaluation, we use a \emph{stratified} and \emph{grouped}
$k$-fold cross-validation where we split the dataset into
$k-1$~challenges for training and 1~challenge for testing. In this
way, we ensure that training is conducted on different challenges than
testing. For each of the $k$~folds, we perform feature selection on
the extracted features and then train the respective classifier as
described in the original publications. We report results averaged
over all 8 folds.

\begin{table}[b]
	\centering
	\footnotesize
	\renewcommand{\arraystretch}{1.1}%
	\begin{tabular}{lccrc}
		\toprule
		\head{Method} &  \head{Lex} & \head{Syn} &
		\head{Classifier} & \head{Accuracy} \\
		\midrule
		\citet{CalHarLiuNar+15} & $\bullet$ &
		$\bullet$ & RF & $90.4\% \; \pm \; 1.7\%$ \\
		\citet{AbuAbuMoh+18} & $\bullet$ &  & LSTM
		& $88.4\% \; \pm \; 3.7\%$  \\
		\bottomrule
	\end{tabular}
	\caption{Implemented attribution methods and their reproduced 
		accuracy. (Lex = Lexical features, Syn~=~Syntactic features)}
	\label{tab:implementation}
\end{table}

\paragraph{Implementation.}  We implement the attribution methods and
our attack on top of Clang~\citep{clang18}, an open-source C/C++
frontend for the LLVM compiler framework. For the method of
\citet{CalHarLiuNar+15}, we re-implement the AST extraction and use
the proposed random forest classifier for attributing programmers. The
approach by \citet{AbuAbuMoh+18} uses lexical
features that are passed to a long short-term memory~(LSTM) neural
network for attribution.  Table~\ref{tab:implementation} provides an
overview of both methods. For further details on the feature
extraction and learning process, we refer the reader to the respective
publications \cite{CalHarLiuNar+15, AbuAbuMoh+18}.

As a sanity check, we reproduce the experiments conducted by
Caliskan et al.~\citep{CalHarLiuNar+15} and \citet{AbuAbuMoh+18} on our
dataset.  Table~\ref{tab:implementation} shows the average attribution
accuracy and standard deviation over the 8~folds.  Our
re-implementation enables us to differentiate the 204~developers with
an accuracy of 90\% and 88\% on average, respectively. Both accuracies
come close to the reported results with a difference of less than
6\%, which we attribute to omitted layout features and the 
stricter dataset. %

\subsection{Untargeted Attack}
\label{sec:untargeted-evasion}

In our first experiment, we investigate whether an adversary can
manipulate source code such that the original author is not
identified. To this end, we apply our untargeted attack to each
correctly classified developer from the 204~authors. We repeat the
attack for all 8~challenges and aggregate the~results.

\paragraph{Attack performance.}
Table~\ref{tab:evasion_summary} presents the performance of the attack
as the ratio of successful evasion attempts. Our attack has a strong
impact on both methods and misleads the attribution in 99\% of the
cases, irrespective of the considered features and learning
algorithm. As a result, the source code of almost all authors can be
manipulated such that the attribution fails.

\begin{table}[t]
	\centering
	\footnotesize
\renewcommand{\arraystretch}{1.1}%
\begin{tabularx}{0.965\linewidth}{lr|rr}
		\toprule
		&  \multicolumn{3}{c}{\head{Success rate of our attack}} \\
	    \head{Method} & \head{Untargeted} &
	    \head{Targeted T+} & \head{Targeted T-} \\
		\midrule
		\citet{CalHarLiuNar+15} &  99.2\% & 77.3\% & 71.2\% \\
		\citet{AbuAbuMoh+18} & 99.1\% & 81.3\% & 69.1\% \\
		\bottomrule
	\end{tabularx}
	\vspace{-0.6em}
	\caption{Performance of our attack as average success rate.
          The targeted attack is conducted with template (T+) and
          without template (T-).}
	\label{tab:evasion_summary}
\end{table}

\begin{figure}[]
	\centering
\includegraphics[]{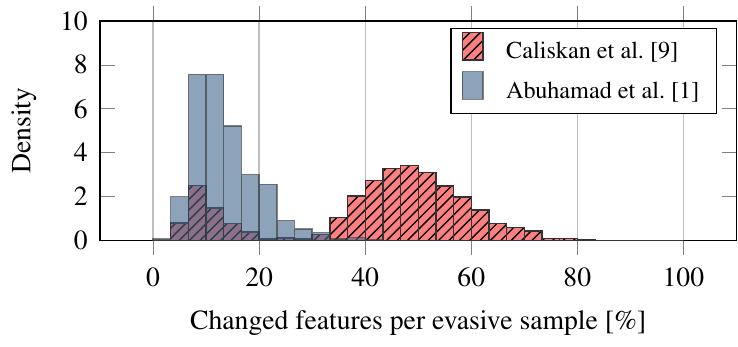}
	\vspace{-0.4em}
	\caption{Untargeted attack: Histogram over the number of changed 
		features per
		successful evasive sample for both attribution methods.}
	\label{fig:dodging_changedfeatures}
\end{figure}

\paragraph{Attack analysis.}
To investigate the effect of our attack in more detail, we compute the
ratio of changed features per adversarial sample.
Figure~\ref{fig:dodging_changedfeatures} depicts the distribution over
all samples. The method by~\citet{CalHarLiuNar+15} exhibits a bimodal
distribution. The left peak shows that a few changes, such as
the addition of include statements, are often sufficient to mislead 
attribution. For the majority of samples, however, the attack alters 
50\% of the features, which indicates the tight correlation between
different features (see Section~\ref{sec:challenges}). A key factor to
this correlation is the TF-IDF weighting that distributes minor
changes over a large set of features.

\begin{figure}
	\centering
\includegraphics[]{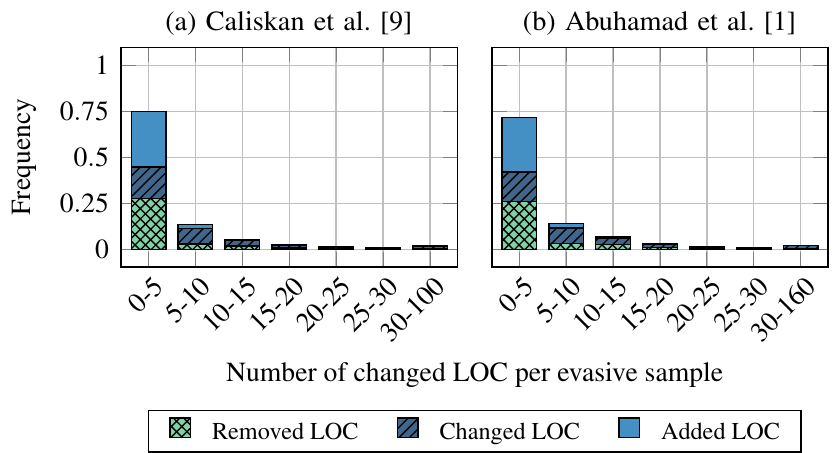}
	\vspace{-1.5em}
	\caption{Untargeted attack: Stacked histogram over the number of 
		changed lines of code
		(LOC) per successful evasive sample for both attribution 
		methods. The original source files have 74 lines on average 
		(std: 38.44).}
	\label{fig:dodging_locchanged}
\end{figure}

In comparison, less features are necessary to evade the approach by
\citet{AbuAbuMoh+18}, possibly due to the higher sparsity of the
feature vectors. Each author has 12.11\% non-zero features on average,
while 53.12\% are set for the method by \citet{CalHarLiuNar+15}. Thus,
less features need to be changed and in consequence each changed
feature impacts fewer other features that remain zero.

Although we observe a high number of changed features, the
corresponding changes to the source code are minimal.
Figure~\ref{fig:dodging_locchanged} shows the number of added, changed
and removed lines of code (LOC) determined by a context-diff with
difflib for each source file before and after the attack. For the
majority of cases in both attribution methods, less than
5~lines of code are added, removed or changed. This low
number exemplifies the targeted design of
our code transformations that selectively alter characteristics of
stylistic patterns.

\paragraph{Summary.}
Based on the results from this experiment, we summarize that our
untargeted attack severely impacts the performance of the methods by
\citet{CalHarLiuNar+15} and \citet{AbuAbuMoh+18}. We conclude that
other attribution methods employing similar features and learning
algorithms also suffer from this problem and hence cannot provide a
reliable attribution in presence of an adversary.

\begin{figure*}
	\begin{subfigure}[b]{.5\linewidth}
		\captionsetup{font=scriptsize}
		\centering
		\includegraphics[width=.90045\columnwidth]{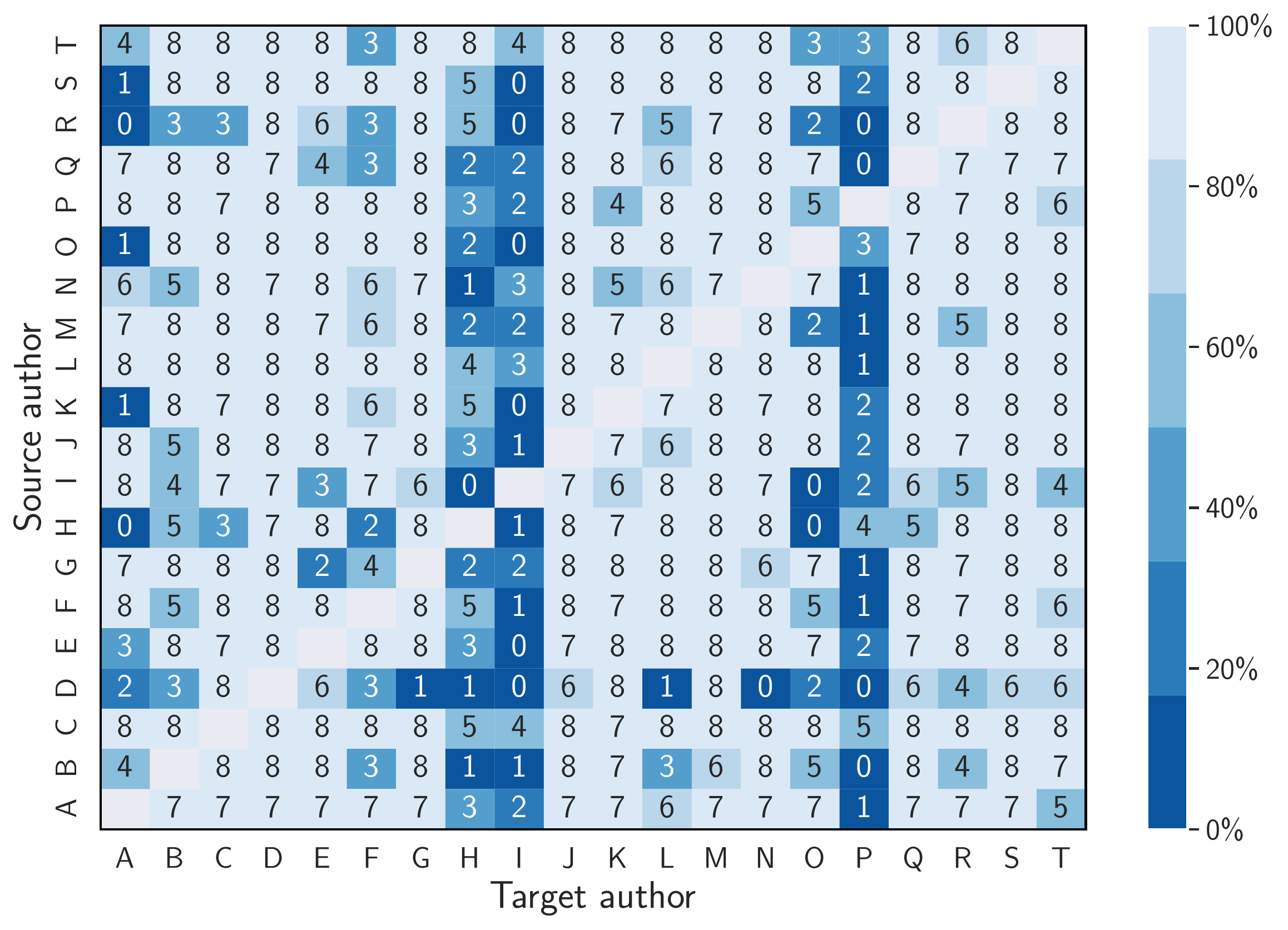}
		\caption{\citet{CalHarLiuNar+15}}\label{fig:evasion_matrixA}
	\end{subfigure}%
	\begin{subfigure}[b]{.5\linewidth}
		\captionsetup{font=scriptsize}
		\centering
		\includegraphics[width=.90045\columnwidth]{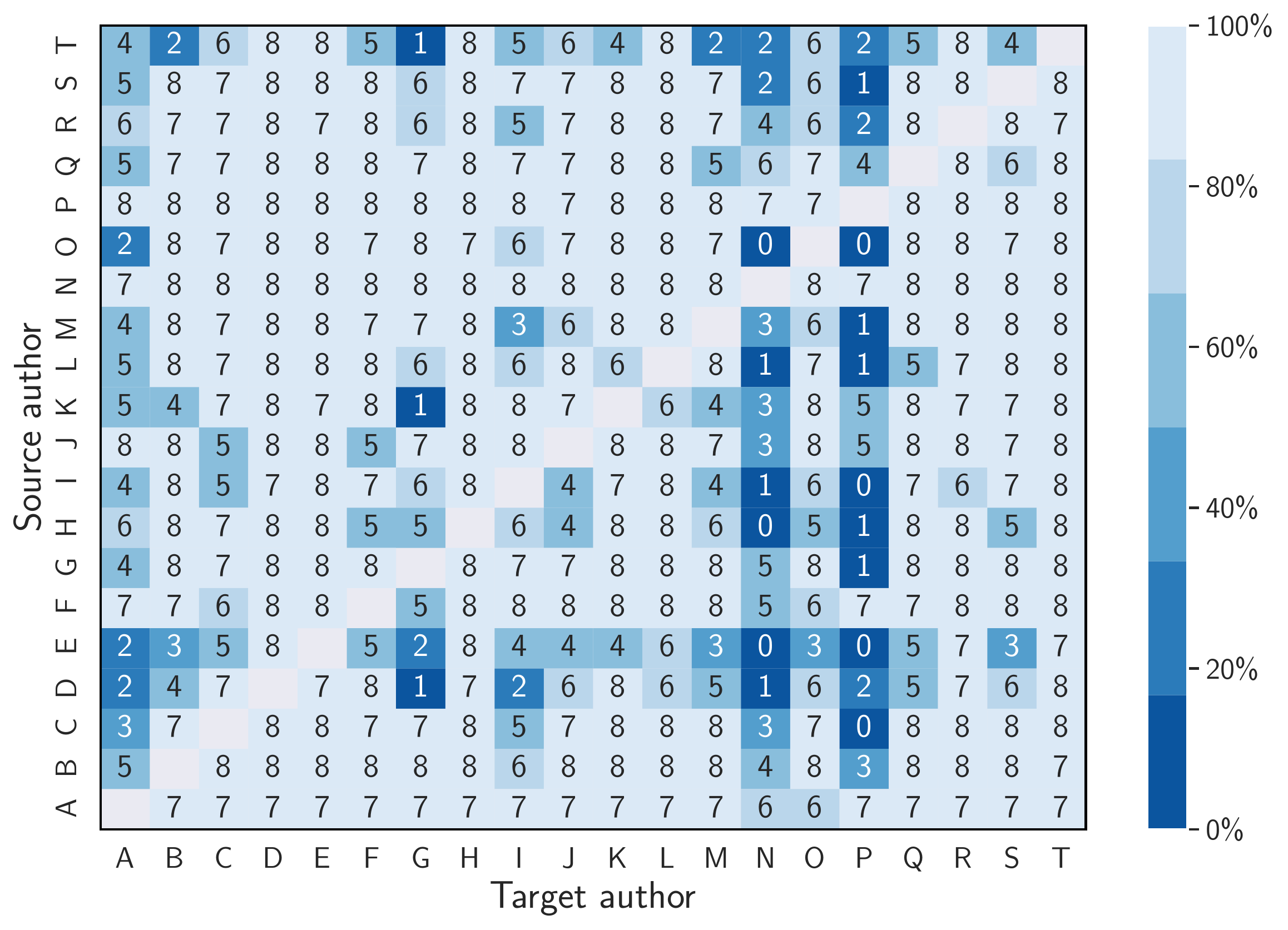}
		\caption{\citet{AbuAbuMoh+18}}\label{fig:evasion_matrixB}
	\end{subfigure}
	\vspace{-0.6cm}
	\caption{Impersonation matrix for both
		attribution methods. Each cell indicates
		the number of successful attack attempts for the
		8~challenges.}\label{fig:evasion_matrices}
\end{figure*}

\subsection{Targeted Attack}
\label{sec:targeted-evasion}

We proceed to study the targeted variant of our attack. We consider 
pairs of programmers, where the
code of the source author is transformed until it is attributed to the
target author. Due to the quadratic number of pairs, we perform this
experiment on a random sample of 20 programmers. This results in
380~source-target pairs each covering the source code of
8~challenges. Table~\ref{tab:list-of-impersonation-authors} in
Appendix~\ref{sec:appendixC} provides a list of the selected authors.
We start with the scenario where we retrieve two samples of
source code for each of the 20 programmers from various GCJ 
challenges---not part of the fixed 8~train-test challenges---to 
support the template transformations.

\paragraph{Attack performance.}
Table~\ref{tab:evasion_summary} depicts the success rate of our attack
for both attribution methods.  We can transfer the prediction from one
to another developer in 77\% and 81\% of all cases, respectively,
indicating that more than three out of four programmers can be
successfully impersonated.

In addition, Figure~\ref{fig:evasion_matrices} presents the results as
a matrix, where the number of successful impersonations is visually
depicted. Note that the value in each cell indicates the absolute
number of successful impersonations for the 8~challenges associated with
each author pair.  We find that a large set of developers can be
imitated by almost every other developer. Their stylistic patterns are
well reflected by our transformers and thus can be easily forged.  By
contrast, only the developers I and P have a small impersonation
rate for \citet{CalHarLiuNar+15}, yet 68\% and 79\% of the developers 
can still imitate the style of I and P in at least one challenge.

\begin{figure}[]
	\centering
\includegraphics[]{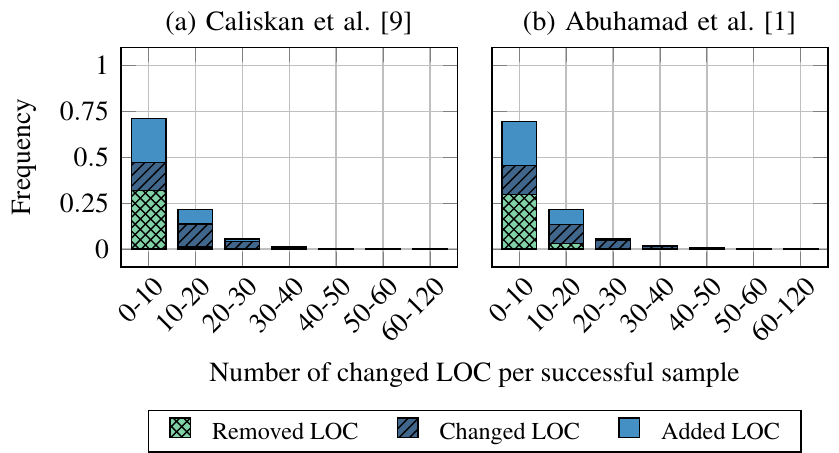}
	\vspace{-1.5em}
	\caption{Targeted attack: Stacked histogram over the number of 
		changed lines of code
		(LOC) per successful impersonation for both attribution 
		methods. The original source files have 74 lines on average 
		(std: 38.44).
	}
	\label{fig:impersonation_locchanged}
\end{figure}

\begin{figure}[]
	\centering
\includegraphics[]{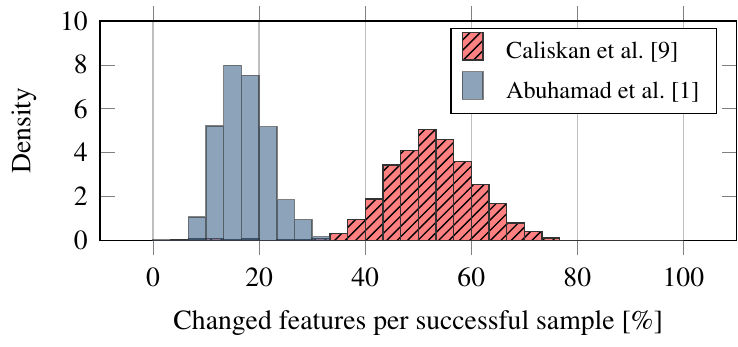}
	\vspace{-.4em}
	\caption{Targeted attack: Histogram over the number of changed 
		features per successful impersonation for both attribution 
		methods.}
	\label{fig:impersonation_changedfeatures}
\end{figure}

\paragraph{Attack analysis.}
The number of altered lines of code also remains small for the
targeted attacks.  Figure~\ref{fig:impersonation_locchanged} shows
that in most cases only 0~to~10~lines of code are affected.  At the
same time, the feature space is substantially changed. 
Figure~\ref{fig:impersonation_changedfeatures} depicts that both
attribution methods exhibit a similar distribution as before in the
untargeted attack---except that the left peak vanishes for the method of
\citet{CalHarLiuNar+15}. This means that each source file requires
more than a few targeted changes to achieve an impersonation.

\begin{table}[h]
	\centering%
	\footnotesize
	\renewcommand{\arraystretch}{1.1}%
	\caption{Usage of transformation families for impersonation}
	\label{tab:code-transformations-imperson}
	\begin{tabular}{p{4.5cm}rr}
		\toprule
		\head{Transformation Family} &
		\head{Cal.~\citep{CalHarLiuNar+15}} &
		\head{Abu.~\citep{AbuAbuMoh+18}} \\
		\midrule
		Control Transformers   & 8.43\% & 9.72\% \\
		Declaration Transformers     & 14.11\% & 17.88\% \\
		API Transformers & 29.90\% & 19.60\% \\
		Miscellaneous Transformers         & 9.15\% & 4.76\% \\
		Template Transformers         & 38.42\% & 48.04\% \\
		\bottomrule
	\end{tabular}
\end{table}

\begin{figure*}[h]
	\centering
	\lstset {
		basicstyle=\ttfamily\scriptsize,
		numbers=left,
		framexleftmargin=2pt,
	}
	\captionsetup[lstlisting]{skip=0pt,labelformat=empty}
	\noindent\begin{minipage}[t]{.44\textwidth}
		\captionof{lstlisting}[ListingsCaption]{Source Author I}
		\begin{lstlisting}[
		]{NameC1}
cout << std::fixed;


for (long long ccr = 1; ccr <= t; ++ccr) {
	double d, n, ans = INT_MIN;
	cin >> d >> n;
	for (double i = 0; i < n; ++i) {
		double k, s;
		cin >> k >> s;
		[...]
	}
	
	ans = d / ans;
	cout << "Case #" << ccr << ": " << setprecision(7) << ans << "\n";

}
		\end{lstlisting}
	\end{minipage}		\hspace{15pt}
	\begin{minipage}[t]{.44\textwidth}
		\captionsetup[lstlisting]{skip=0pt,labelformat=empty}
		\captionof{lstlisting}[ListingsCaption]{Source 
		$\longrightarrow$ Target}
		\begin{lstlisting}[
		escapechar=~,
		]{NameC2}
typedef double td_d; ~\tikz[remember picture] \node [] (CA) {};~
[...]
long long ccr = 1;  ~\tikz[remember picture] \node [] (CBC) {};~
while (ccr <= t) {   ~\tikz[remember picture] \node [] (CB) {};~
	td_d d, n, ans = INT_MIN;
	cin >> d >> n;
	td_d i; ~\tikz[remember picture] \node [] (CC) {};~
	for (i = 0; i < n; ++i) { 
		td_d k, s; ~\tikz[remember picture] \node [] (CAA) {};~
		cin >> k >> s;
		[...]
	}
	ans = d / ans;
	printf("Case #%lld: %.7f\n", 
		ccr, ans); ~\tikz[remember picture] \node [] (CD) {};~
	++ccr; ~\tikz[remember picture] \node [] (CBB) {};~
}
\end{lstlisting}
\begin{tikzpicture}[remember picture, overlay,
	font=\scriptsize,
	ar/.style = {dashed, sunygreen!50!black, thick},
	arr/.style = {->, ar},
	]
	\node at (7.25,0) (d){};

	% Draw nodes with number	
	\draw[fill=red!30, draw] (d |- CA)
	node[ar] {\normalsize{\ding{182}}};
	\draw[fill=red!30, draw] (d |- CB) 
	node[ar] {\normalsize{\ding{183}}};
	\draw[fill=red!30, draw] (d |- CC) 
	node[ar,yshift=-3.75] {\normalsize{\ding{184}}};
	\draw[fill=red!30, draw] (d |- CD)
	node[ar, yshift=3.75pt] {\normalsize{\ding{185}}};

% Make arrows for for->while transformer:
\coordinate (wh) at (4,0); % common x-value for CBB and CBC
\coordinate (While) at ($(d |- CB) + (-1.25, 0)$);
\coordinate (almostCBNode) at ($(d |- CB) + (-0.375,0)$);
\draw[ar] (wh |- CBB) -| ([yshift=-2.5]While) -- 
							([yshift=-2.5]almostCBNode);
\draw[ar] (wh |- CBC) -| ([yshift=2.5]While) -- 
							([yshift=2.5]almostCBNode);

\draw[ar] (wh |- CB) -- (almostCBNode);

\end{tikzpicture} 
\end{minipage}

\begin{minipage}[T]{.44\textwidth}
		\captionsetup[lstlisting]{skip=0pt,labelformat=empty}
		\captionof{lstlisting}[ListingsCaption]{Target Author P}
		\begin{lstlisting}[
		]{NameC3}
int T, cas = 0;
cin >> T;
while (T--) {
  int d, n;
  cin >> d >> n;
  double t = 0;
  while (n--) {
	  int k, s;
	  cin >> k >> s;
	  t = max((1.0 * d - k) / s, t);
  }
  double ans = d / t;
  printf("Case #%d: %.10f\n", ++cas, ans);
}
		\end{lstlisting}
              \end{minipage} 	\hspace{15pt}
\begin{minipage}[T]{.44\textwidth}
  \vspace{-0.6em} \renewcommand{\arraystretch}{1.5} \scriptsize
  \centering
             \begin{tabular}{clp{43mm}}
              \head{Iteration} & \head{Transformer} & \head{Description} \\
              \midrule
              \textcolor{sunygreen!50!black}{\normalsize{\ding{182}}}
              & Typedef & adds typedef and replaces all locations 
              with previous type by novel typedef. \\
              \textcolor{sunygreen!50!black}{\normalsize{\ding{183}}}
              & For statement & converts \code{for}-statement into an 
              equivalent \code{while}-statement, as target tends to 
              solve problems via \code{while}-loops. \\
              \textcolor{sunygreen!50!black}{\normalsize{\ding{184}}}
              & Init-Decl & moves a declaration out of the control 
              statement which mimics the declaration behavior of 
              while-statements.\\
              \textcolor{sunygreen!50!black}{\normalsize{\ding{185}}}
              & Output API & 
              substitutes C++ API for writing output by 
              C API \code{printf}. To this end, it determines the 
              precision of output statements by finding \code{fixed} 
              (line~1) and \code{setprecision} (line~14) commands.
              \\
               \end{tabular}

          \end{minipage}
	\vspace{-0.8em}
	\caption{Impersonation example from our evaluation for the GCJ 
	problem \emph{Steed 2: Cruise Control}. The upper left listing
	shows the original source file, the upper right its 
	modified version such that it is classified as the target author.
	For comparison, the lower left listing shows the original source 
	file from the target author (which was not available for the 
	attacker). The table lists the necessary transformations.
		}
	\label{fig:eval_code_compare}
\end{figure*}

Table~\ref{tab:code-transformations-imperson} shows the contribution
of each transformation family to the impersonation success. All
transformations are necessary to achieve the reported attack rates.  A
closer look reveals that the method by \citet{AbuAbuMoh+18} strongly
rests on the usage of template transformers, while the families are
more balanced for the approach by  \citet{CalHarLiuNar+15}. This
difference can be attributed to the feature sets, where the
former method relies on simple lexical features only and the latter
extracts more involved features from the~AST.

\paragraph{Case Study.}
To provide further intuition for a successful impersonation,
Figure~\ref{fig:eval_code_compare} shows a case study from our
evaluation. The upper two panels present the code from the source
author in original and transformed form.  The lower left panel depicts
the original source text from the target author for the same
challenge. Note that the attack has no access to this file. The table
lists four~conducted transformations. For instance, the target author 
has the stylistic pattern to use \code{while} statements, C functions 
for the output, and particular typedefs. By changing these patterns, our
attack succeeds in misleading the attribution
method. %

\paragraph{Attack without template.}
We additionally examine the scenario when the adversary has no access
to a template file of the target developer. In this case, our template 
transformers can only try common patterns, such as the
iteration variables \code{i}, \code{j}, \dots, \code{k} or typedef 
\code{ll} for the type \mbox{\code{long long}}. 
Table~\ref{tab:evasion_summary}
shows the results of this experiment as well. Still, we achieve an
impersonation rate of~71\% and 69\%---solely by relying on the
feedback from the classifier. The number of altered lines of code and 
features correspond to Figures~\ref{fig:impersonation_locchanged} 
and~\ref{fig:impersonation_changedfeatures}.

Contrary to expectation, without a template, the approach by 
\mbox{\citet{AbuAbuMoh+18}} is harder
to fool than the method by \citet{CalHarLiuNar+15}. 
As the lexical features rest more on simple
declaration names and included libraries, they are harder to guess
without a template file. However, if a template file is available,
this approach is considerably easier to evade.

\paragraph{Attack with substitute model.} 
\label{subsec:substituteexperiments}
We finally demonstrate that an impersonation is even possible without 
access to the prediction of the original classifier, only relying on a 
substitute model trained from separate data.
We split our training set into disjoint sets with three files 
per author to train the original and substitute model, respectively.
We test the attack on the method by \citet{CalHarLiuNar+15}, which is 
the more robust attribution under attack. %
By the nature of this scenario, the adversary can use two 
files to support the template transformations.

Adversarial examples---generated with the substitute 
model---transfer in 79\% of the cases to the original model,
that is, attacks successful against the substitute model are also 
effective against the original in the majority of the cases.
This indicates that our attack successfully changes indicative features
for a target developer across models. 
The success rate of our attack on the original 
model is 52\%. Due to the reduced number of training files in this 
experiment, the attack is harder, as the coding habits are less 
precisely covered by the original and substitute models. 
Still, we are able to impersonate every second developer with no 
access to the original classifier.

\paragraph{Summary.}
Our findings show that an adversary can automatically impersonate a
large set of developers without and with access to a template file.
We conclude that both considered attribution methods can be abused to
trigger false allegations---rendering a real-world application
dangerous.

\subsection{Preserved Semantics and Plausibility}
\label{sec:pers-semant-plaus}

In the last experiment, we verify that our adversarial code samples
comply with the attack constraints specified in
Section~\ref{sec:attackconstraints}. That is, we empirically check
that (a)~the semantics of the transformed source code are preserved,
(b)~the generated code is plausible to a human analyst, and (c)~layout
features can be trivially evaded.

\paragraph{Preserved semantics.} We begin by verifying the semantics
of the transformed source code.  In particular, we use the test file
from each challenge of the GCJ competition to check that the
transformed source code provides the same solution as the original
code. In all our experiments, we can verify that the output remains
unchanged for each manipulated source code sample before and after
our attack.

\paragraph{Plausible code.}
Next, we check that our transformations lead to plausible code and
conduct a discrimination test with 15~human subjects.  The group
consists of 4~undergraduate students, 6~graduate students and
5~professional computer scientists.  The structure of the test follows
an \emph{AXY-test}: Every participant obtains 9~files of source
code---each from a different author but for the same GCJ challenge.
These 9~files consists of 3~unmodified source codes as reference (A)
and 6~sources codes (XY) that need to be classified as either
\emph{original} or \emph{modified}. The participants are
informed that 3~of the samples are modified. We then ask each
participant to identify the unknown samples and to provide a short
justification.

The results of this empirical study are provided in
Table~\ref{tab:study}. On average, the participants are able to
correctly classify 60\% of the provided files which is only marginally
higher than random guessing.  This result highlights that it is hard
to decide whether source code has been modified by our attack or not.
In several cases, the participants falsely assume that unused
\code{typedef} statements or an inconsistent usage of operators are
modifications.
\begin{table}[bht]
	\centering%
		\footnotesize
	\renewcommand{\arraystretch}{1.1}%
	\caption{Study on plausibility of transformed source code.}
	\label{tab:study}
	\begin{tabular}{p{4.50cm}rr}
		\toprule
		\head{Participant Group} & \head{Accuracy} &\head{Std} \\
		\midrule
		Undergraduate students & 66.7\% & 23.6\% \\
		Graduate students & 55.6\% & 15.7\% \\
		Professionals & 60.0\% & 24.9\% \\\midrule
		Total  & 60.0\% & 21.8\% \\
		Random guessing & 50.0\% & --- \\
		\bottomrule
	\end{tabular}
\end{table}

\paragraph{Evasion of layout features.}
Finally, we demonstrate that layout features can be trivially
manipulated, so that it is valid to restrict our approach to the
forgery of lexical and syntactic features. To this end, we train a
random forest classifier \emph{only} on layout features as extracted
by \citet{CalHarLiuNar+15}. We then compare the
attribution accuracy of the classifier on the test set with and
without the application of the formatting tool \code{clang-format},
which normalizes the layout of the code.

While the attribution method can identify 27.5\% of the
programmers based on layout features if the code is not formatted, the
performance decreases to 4.5\% if we apply the formatting tool to the
source code.  We thus conclude that it is trivial to mislead an
attribution based on layout features.

\section{Limitations}
\label{sec:limitations}

Our previous experiments demonstrate the impact of our attack on
program authorship attribution. Nonetheless, our approach has
limitations which we discuss in the following.

\paragraph{Adversarial examples $\boldsymbol\neq$ anonymization.} Our
attack enables a programmer to hide their identity in source code by
misleading an
attribution. While such an attack protects the privacy of the
programmer, it is not sufficient for achieving anonymity. Note that
$k$-anonymity would require a set of $k$~developers that are equally
likely to be attributed to the source code. In our setting, the code
of the programmer is transformed to match a different author and an
anonymity set of sufficient size is not guaranteed to exist. Still, we
consider anonymization as promising direction for further research,
which can build on the concepts of code transformations developed in
this paper.

\paragraph{Verification of semantics.} Finally, we consider two
programs to be semantically equivalent if they return the same output
for a given input. In particular, we verify that the transformed
source code is semantically equivalent by applying
the test cases provided by the GCJ competition. Although this approach
is reasonable in our setting, it cannot guarantee strict semantic
equivalence in all possible cases. Some of the exchanged API
functions, for example, provide the same functionality but differ in
corner cases, such as when the memory is exhausted. We acknowledge
this limitation, yet it does not impact the general validity of our
results.

\section{Related Work}
\label{sec:related-work}

The automatic attack of source-code authorship attribution touches
different areas of security research. In this section, we review
related methods and concepts.

\paragraph{Authorship attribution of source code.}
Identifying the author of a program is a challenging task of computer
security that has attracted a large body of work in the last
years. Starting from early approaches experimenting with hand-crafted
features \citep[][]{MacGraMacSal99,KrsSpa97}, the techniques for
examining source code have constantly advanced, for example, by
incorporating expressive features, such as n-grams
\citep[e.g.,][]{FraStaGri+06, BurUitTur09, AbuAbuMoh+18} and abstract
syntax trees
\citep[e.g.,][]{Pel00, CalHarLiuNar+15, AlsDauHarMan+17}.  Similarly,
techniques for analyzing native code and identifying authors of
compiled programs have advanced in the last years
\citep[e.g.,][]{AlrShiWanDeb+18, CalYamTau+18, MenMilJun17,
	RosZhuMil11}.

Two notable examples for source code are the approach by
Caliskan et al.~\cite{CalHarLiuNar+15} and by \citet{AbuAbuMoh+18}.
The former inspects features derived from code
layout, lexical analysis and syntactic analysis. Regarding
comprehensiveness, this work can be considered as the current state of
the art. The work by \citet{AbuAbuMoh+18} focuses on lexical features
as input for recurrent neural networks. Their work covers the largest
set of authors so far and makes use of recent advances in deep learning.
Table~\ref{tab:related} shows the related approaches.

\begin{table}[htbp]
	\centering
			\footnotesize
	\begin{tabular}{lcccrr}
		\toprule
		\bf Method & \bf Lay & \bf Lex & \bf Syn & \bf Authors & \bf
		Results \\
		\midrule
		*\textbf{\citet{AbuAbuMoh+18}} &  & $\bullet$ &  &
		\textbf{8903} & \textbf{92\%} \\
		*\textbf{\citet{CalHarLiuNar+15}} & $\bullet$ & $\bullet$ &
		$\bullet$ & \textbf{250} & \textbf{95\%} \\
		\citet{AlsDauHarMan+17} &  &  & $\bullet$ & 70 & 89\% \\
		\citet{FraStaGri+06} & $\bullet$ & $\bullet$ & & 30 & 97\% \\
		\citet{KrsSpa97} & $\bullet$ & $\bullet$ & $\bullet$ & 29 &
		73\% \\
		\citet{BurUitTur09}  &  $\bullet$ &  $\bullet$ &  & 10 & 77\% \\
		\bottomrule
	\end{tabular}
	\caption{Comparison of approaches for source code authorship
	attribution.
		Lay = Layout features, Lex = Lexical features, Syn = Syntactic
		features. *Attacked in this paper.}
	\label{tab:related}
\end{table}

Previous work, however, has mostly ignored the problem of untargeted and
targeted attacks.  Only the empirical study by 
\mbox{\citet{SimZetKoh18}}
examines how programmers can mislead the attribution by
\citet{CalHarLiuNar+15} by mimicking the style of other
developers. While this study provides valuable insights into the risk
of forgeries, it does not consider automatic attacks and thus is
limited to manipulations by humans.  In this paper, we demonstrate
that such attacks can be fully automated. Our generated forgeries even
provide a higher success rate than the handcrafted samples in the
study.  Moreover, we evaluate the impact of different feature sets and
learning algorithms by evaluating two attribution methods.

\paragraph{Adversarial machine learning.}
The security of machine learning techniques has also attracted a lot
of research recently. A significant fraction of work on attacks has
focused on scenarios where the problem and feature space are mainly
identical~\citep[see][]{BigCorMai+13, PapMcDJhaFreCelSwa16,
  CarWag17}. In these scenarios, changes in the problem space, such as
the modification of an image pixel, have a one-to-one effect on the
feature space, such that sophisticated attack strategies can be
applied. By contrast, a one-to-one mapping between source code and the
extracted features cannot be constructed and thus we are required to
introduce a mixed attack strategy (see Section~\ref{sec:attack}).

Creating evasive PDF malware samples~\citep{RndLas14, XuQiEva16} and 
adversarial examples for text classifiers~\citep[e.g.,][]{AlzShaElg+18, 
LiJiDu+19} represent two similar scenarios, where the practical 
feasibility needs to be ensured. These works typically operate in the 
problem space, where search algorithms such as hill climbing or genetic 
programming are guided by information from the feature space.
MCTS represents a novel concept in the portfolio of creating 
adversarial examples under feasibility constraints, previously examined 
by \citet{WicHuaKwi18} in the image context only.

Also related is the approach by \citet{ShaBhaBauRei16} for misleading
face recognition systems using painted eyeglasses. The proposed attack
operates in the feature space but ensures practical feasibility by
refining the optimization problem. In particular, the calculated
adversarial perturbations are required to match the form of
eyeglasses, to be printable, and to be invariant to slight head
movements. In our attack scenario, such refinements of the
optimization problem are not sufficient for obtaining valid source
code, and thus we resort to applying code transformations in the
problem space.

\section{Conclusion}
\label{sec:conclusion}
Authorship attribution of source code can be a powerful tool if an
accurate and robust identification of programmers is possible.  In
this paper, however, we show that the current state of the art is
insufficient for achieving a robust attribution.  We present a
black-box attack that seeks adversarial examples in the domain of
source code by combining Monte-Carlo tree search with concepts from
source-to-source compilation. Our empirical evaluation shows that
automatic untargeted and targeted attacks are technically feasible and
successfully mislead recent attribution methods.

Our findings indicate a need for alternative techniques for
constructing attribution methods. These techniques should be designed
with robustness in mind, such that it becomes harder to transfer
stylistic patterns from one source code to another.  A promising
direction are generative approaches of machine learning, such as
generative adversarial networks, that learn a decision function while
actively searching for its weak spots. Similarly, it would help to
systematically seek for stylistic patterns that are inherently hard to
manipulate, either due to their complexity or due to their tight
coupling with program semantics.

\paragraph{Public dataset and implementation.} To encourage further
research
on program authorship attribution and, in particular, the development of
robust methods, we make our dataset and implementation publicly
available.\footnote{www.tu-braunschweig.de/sec/research/code/imitator}
The attribution
methods, the code transformers as well as our attack algorithm are
all implemented as
individual modules, such that they can be easily combined and extended.

\section*{Acknowledgment}
The authors would like to thank Johannes Heidtmann for his 
assistance during the project, and the
anonymous reviewers for their suggestions and comments.
Furthermore, the authors acknowledge funding by the Deutsche 
Forschungsgemeinschaft 
(DFG, German Research Foundation) under Germany's Excellence Strategy 
- \mbox{EXC 2092 CASA - 390781972} and the research grant \mbox{RI 
2469/3-1}.

\bibliographystyle{abbrvnat}

\appendix

\section{Monte-Carlo Tree Search}\label{sec:appendixE}

In this section, we provide further details about our variant of 
\emph{\mbox{Monte-Carlo tree search}}. 
Algorithm~\ref{alg:alg_mcts} gives an overview of the attack.
The procedure \codeintext{Attack} starts with the root node~$r_0$ that
represents the original source code $x$. The algorithm then works in 
two nested loops:
\begin{itemize}
	\item 
	The outer loop in lines 3--5 repetitively builds a search 
	tree for the current state of source code $r$, and takes a single 
	move (i.e. a single transformation). To do so, in each iteration, 
	we choose the child node with the highest average classifier score.
	This process is repeated until the attack succeeds or a 
	stop criterion is fulfilled (we reach a fixed number of outer 
	iterations or we do not observe any improvement over multiple 
	iterations)~(line~3).
	\item 
	The procedure \codeintext{MCTS} represents the inner loop. 
	It iteratively builds and extends the search tree under the current 
	root node $r$. As this procedure is the main building block
	of our attack, we discuss the individual steps in more 
	detail in the following.
\end{itemize}
\vspace{-0.2em}

\begin{algorithm}
	\footnotesize
	\caption{Monte-Carlo Tree Search}\label{alg:alg_mcts}
	\begin{algorithmic}[1]
		\Procedure{Attack}{$r_0$}
		\State $r \gets r_0$
		\While{not \Call{Success}{$r$} and not 
		\Call{StopCriterion}{$r$}}
		\State \Call{MCTS}{$r$}
		\Comment{Extend the search tree under $r$}
		\State $r \gets $ \Call{ChildWithBestScore}{$r$}
		\Comment{Perform next move}
		\EndWhile\label{euclidendwhile}
		\EndProcedure
		\Procedure{MCTS}{$r$}
		\For{$i\gets 1, N$}
		\State $u \gets$ \Call{Selection}{$r$, $i$}
		\State $\mathcal{T} \gets$ \Call{Simulations}{$u$}
		\State \Call{Expansion}{$u$, $\mathcal{T}$}
		\State \Call{Backpropagation}{$\mathcal{T}$}
		\EndFor
		\EndProcedure
	\end{algorithmic}
\end{algorithm}

\paragraph{Selection.}
Algorithm~\ref{alg:alg_mcts_selection} shows the pseudocode to find the
next node which is evaluated. The procedure recursively selects a child 
node according to a selection policy. We stop if the
current node has no child nodes or if we have not marked it before in
the current procedure \codeintext{Selection}. 
The procedure finally returns the node that will be evaluated 
next.

As the number of possible paths grows exponentially (we have up to
36~transformations as choice at each node), we cannot evaluate all
possible paths. The tree creation thus crucially depends on a
selection policy. We use a simple heuristic to approximate the 
\emph{Upper Confidence Bound for Trees} algorithm that is often used as 
selection policy 
(see~\citep{BroPowWhi+12}). Depending on the current iteration index 
$i$ of \codeintext{Selection}, the procedure 
\codeintext{SelectionPolicy} 
alternately returns the decision rule to choose the child with the 
highest average score, the lowest visit count or the highest score 
standard deviation. This step balances the 
\emph{exploration} of less-visited nodes and the \emph{exploitation} of 
promising nodes with a high average score.

\begin{algorithm}
	\footnotesize
	\caption{Selection Procedure of MCTS}\label{alg:alg_mcts_selection}
	\begin{algorithmic}[1]
		\Procedure{Selection}{$r$, $i$}
		\State $D \gets$ \Call{SelectionPolicy}{$i$}
		\State $u \gets r$
		\While{$u$ has child nodes}
		\State $v \gets \Call{SelectChild}{u, D}$
		\Comment{Child of $u$ w.r.t. to $D$}
		\If{$v$ not marked as visited}
		\State Mark $v$ as visited
		\State \Return{$v$}
		\Else
		\State $u \gets v$
		\EndIf
		\EndWhile
		\EndProcedure
	\end{algorithmic}
\end{algorithm}

\paragraph{Simulations.}
Equipped with the node $u$ that needs to be evaluated, the next
step generates a set of transformation sequences
$\mathcal{T}$ that start at~$u$:
\begin{align}
\mathcal{T} = \lbrace \mathbf{T}_j \; \vert j = 1,\dots,k \;
\text{ and } \vert \mathbf{T}_j \vert \leq M \rbrace \; ,
\end{align}
where $|\mathbf{T}_j|$ is the number of transformations in 
$\mathbf{T}_j$.
The sequences are created randomly and have a varying length which is,
however, limited by $M$. In our experiments, we set $M=5$ to reduce the 
number of possible branches. 

In contrast to the classic
game use-case, we can use the returned scores $g(x)$ as early
feedback and thus we do not need to play out a full game. In other
words, it is not necessary to evaluate the complete path to obtain
feedback. For each sequence, we determine the classifier score by 
passing the modified source code at the end of each sequence to the 
attribution method.
We further pinpoint a difference to the general MCTS algorithm.
Instead of evaluating only one path, we create a batch of sequences
that can be efficiently executed in parallel. In this way, we reduce
the computation time and obtain the scores for various paths.

\paragraph{Expansion.}
We continue by inserting the respective transformations from the 
sequences as novel tree nodes under $u$ (see 
Algorithm~\ref{alg:alg_mcts_expand_tree}).
For each sequence, we start with $u$ and the first transformation. We 
check if a child node with the same transformation already 
exists under $u$. If not, a new node $v$ is created and added as child 
under $u$. Otherwise, we use the already existing node $v$. We repeat 
this step with $v$ and the next transformation.
Figure~\ref{fig:mcts_figure} from Section~\ref{sec:attackstrategy} 
exemplifies this expansion step.

\begin{algorithm}
	\footnotesize
	\caption{Expansion Procedure of MCTS}
	\label{alg:alg_mcts_expand_tree}
	\begin{algorithmic}[1]
		\Procedure{Expansion}{$u$, $\mathcal{T}$}

		\For{$\mathbf{T}$ in $\mathcal{T}$} \Comment{For each 
		sequence}
		\State $z \gets u$
		\For{$T$ in $\mathbf{T}$} \Comment{For each transformer}
		\If{$z$ has no child with $T$}
			\State $v \gets $ \Call{CreateNewNode}{$T$}
			\State $z$.add\_child($v$)
		\Else
			\State $v \gets $ \Call{$z$.getChildWith}{$T$}
		\EndIf
		\State $z \gets v$
		\EndFor
		\EndFor
		\EndProcedure
	\end{algorithmic}
\end{algorithm}

\paragraph{Backpropagation.}
Algorithm~\ref{alg:alg_mcts_backpropagation} shows the last step that
backpropagates the classifier scores to the root. For each sequence,
the procedure first determines the last node~$n$ of the current
sequence and the observed classifier score $s$ at node $n$.
Next, all nodes on the path from $n$ to the root node of
the search tree are updated. First, the visit count of each path node 
is incremented. Second, the final classifier score~$s$ is added to the 
score list of each path node.
Both statistics are used by \codeintext{SelectChild} to choose the next 
promising node for evaluation. 
Furthermore, \codeintext{ChildWithBestScore} uses the score list to 
obtain the child node with the highest average score.

\begin{algorithm}
	\footnotesize
	\caption{Backpropagation Procedure of MCTS}
	\label{alg:alg_mcts_backpropagation}
	\begin{algorithmic}[1]
		\Procedure{Backpropagation}{$\mathcal{T}$}
\For{$\mathbf{T}$ in $\mathcal{T}$}

\State $s \gets $ \Call{GetScore}{$\mathbf{T}$}
\State get $n$ as tree leaf of current sequence %
\While{$n$ is not None} \Comment{Backpropagate to root}
\State $n$.visitCount $\gets$ $n$.visitCount + 1 \Comment{Increase
visit count}
\State $n$.scores = $n$.scores $\cup$ $s$ \Comment{Append score}
\State $n \gets n$.parent \Comment{Will be None for root node}
\EndWhile

\EndFor
		\EndProcedure
	\end{algorithmic}
\end{algorithm}

We finally note a slight variation for the scenario with a 
substitute model (see Section~\ref{sec:attackobjectives}). To improve 
the transferability rate from the substitute to the original model, we 
do not terminate at the first successful adversarial example. Instead, 
we collect all successful samples and stop the outer loop after a 
predefined number of iterations. We choose the sample with the highest 
score on the substitute to be tested on the original classifier.

\section{List of Developers For Impersonation}\label{sec:appendixC}

Table~\ref{tab:list-of-impersonation-authors} maps the letters to
the 20 randomly selected programmers from the 2017 GCJ contest.

\begin{table}[h]
	\centering%
	\footnotesize
	\renewcommand{\arraystretch}{1.1}%
	\caption{List of developers for impersonation}
	\label{tab:list-of-impersonation-authors}
	\begin{tabular}{lp{2.15cm}p{0.25cm}lp{2.15cm}}
		\toprule
		\head{Letter} & \head{Author} & & \head{Letter} & \head{Author}
		\\
		\midrule
A & 4yn & & K & chocimir \\
B & ACMonster & & L & csegura \\
C & ALOHA.Brcps & & M & eugenus \\
D & Alireza.bh & & N & fragusbot \\
E & DAle & & O & iPeter \\
F & ShayanH & & P & jiian \\
G & SummerDAway & & Q & liymouse \\
H & TungNP & & R & sdya \\
I & aman.chandna & & S & thatprogrammer \\
J & ccsnoopy & & T & vudduu \\
		\bottomrule
	\end{tabular}
\end{table}

\section{List of Code Transformations}\label{sec:appendixA}

A list of all 36~developed code transformations is presented in
Table~\ref{tab:appendix-transf}. The transformers are
grouped accordingly to the family of their implemented
transformations, i.e, transformations altering the control flow,
transformations of declarations, transformations replacing the used
API, template transformations, and miscellaneous transformations.

\begin{table*}[p]
	\centering
	\footnotesize
	\caption{List of Code Transformations}
	\label{tab:appendix-transf}
	\begin{subtable}{\linewidth}
		\begin{tabularx}{\textwidth}{p{.21\textwidth}X}
			\toprule
		\end{tabularx}
		\caption*{\bf Control Transformations}
		\begin{tabularx}{\textwidth}{p{.21\textwidth}X}
			\head{Transformer} & \head{Description of Transformations}
			\\
			\midrule
			For statement transformer & Replaces a \code{for}-statement
			by an equivalent \code{while}-statement. \\
			\addlinespace[2pt]
			While statement transformer & Replaces a
			\code{while}-statement
			by an equivalent \code{for}-statement. \\
			\addlinespace[2pt]
			Function creator & Moves a whole block of code to a
			standalone
			function and creates a call to the new function at the
			respective
			position. The transformer identifies and passes all
			parameters
			required by the new function.
			It also adapts statements that change the control flow
			(e.g.\ the
			block contains a \code{return} statement that also needs to
			be
			back propagated over the caller).
			\\
			\addlinespace[2pt]
			Deepest block transformer & Moves the deepest block in the
			AST to a standalone function.\\
			\addlinespace[2pt]
			If statement transformer & Split the condition of a single
			\code{if}-statement at logical operands (e.g., \code{\&\&}
			or \code{||}) to create a cascade or a sequence of two
			\code{if}-statements depending on the logical operand. \\
		\end{tabularx}
	\end{subtable}
	\newline\bigskip\newline
	\begin{subtable}{\linewidth}
		\caption*{\bf Declaration Transformations}
		\begin{tabularx}{\textwidth}{p{.21\textwidth}X}
			\head{Transformer} & \head{Description of Transformation} \\
			\midrule
			Array transformer & Converts a static or dynamically
			allocated array into a C++ vector object. \\
			\addlinespace[2pt]
			String transformer & \emph{Array option:} Converts a char
			array (C-style string) into a C++ string object. The
			transformer adapts all usages in the respective scope, for
			instance, it replaces all calls to \code{strlen} by calling
			the instance methods \code{size}. \\
			& \emph{String option:}  Converts a C++ string object into
			a char array (C-style string). The transformer adapts all
			usages in the respective scope, for instance, it deletes
			all calls to \code{c\_str()}. \\
			\addlinespace[2pt]
			Integral type transformer & Promotes integral types
			(\code{char}, \code{short}, \code{int}, \code{long},
			\code{long long}) to the next higher type, e.g., \code{int}
			is replaced by \code{long}. \\
			\addlinespace[2pt]
			Floating-point type transformer & Converts \code{float} to
			\code{double} as next higher type.\\			
			\addlinespace[2pt]
			Boolean transformer & \emph{Bool option:} Converts true or
			false by an integer representation to exploit the implicit
			casting. \\
			& \emph{Int option:} Converts an integer type into a
			boolean type if the integer is used as boolean value only.
			\\
			\addlinespace[2pt]
			Typedef transformer & \emph{Convert option:} Convert a type
			from source file to a new type via \code{typedef}, and
			adapt all locations where the new type can be used. \\
			& \emph{Delete option:} Deletes a type definition
			(\code{typedef}) and replace all usages by the original
			data type. \\
			\addlinespace[2pt]
			Include-Remove transformer & Removes includes from source
			file that are not needed.\\
			\addlinespace[2pt]
			Unused code transformer & \emph{Function option:} Removes
			functions that are never called. \\
			& \emph{Variable option:} Removes global variables that are
			never used. \\
			\addlinespace[2pt]
			Init-Decl transformer & \emph{Move into option:}
			Moves a declaration for a control statement if defined
			outside into the control statement. For instance, \code{int
			i; ...; for(i = 0; i < N; i++)} becomes \code{for(int i =
			0; i < N; i++)}.\\
			& \emph{Move out option:} Moves the declaration of a
			control statement's initialization variable out of the
			control statement.
			\\
		\end{tabularx}
	\end{subtable}
	\newline\bigskip\newline
	\begin{subtable}{\linewidth}
		\caption*{\bf API Transformations}
		\begin{tabularx}{\textwidth}{p{.21\textwidth}X}
			\head{Transformer} & \head{Description of Transformations}
			\\
			\midrule
			Input interface transformer & \emph{Stdin option:} Instead
			of
			reading the input from a file (e.g. by using the API
			\code{ifstream} or \code{freopen}), the input to the
			program is read from stdin directly (e.g. \code{cin} or
			\code{scanf}). \\
			& \emph{File option:} Instead of reading the input from
			stdin, the input is retrieved from a file. \\
			\addlinespace[2pt]
			Output interface transformer & \emph{Stdout option:}
			Instead of printing the output to a file (e.g. by
			\code{ofstream} or \code{freopen}), the output is written
			directly to stdout (e.g. \code{cout} or \code{printf}). \\
			& \emph{File option:} Instead of writing the output
			directly to stdout, the output is written to a file. \\
			\addlinespace[2pt]
			Input API transformer & \emph{C++-Style option:}
			Substitutes C APIs used for reading input (e.g.,
			\code{scanf}) by C++ APIs (e.g., usage of \code{cin}). \\
			& \emph{C-Style option:} Substitutes C++ APIs used for
			reading input (e.g., usage of \code{cin}) by C APIs (e.g.,
			\code{scanf}). \\
			\addlinespace[2pt]
			Output API transformer & \emph{C++-Style option:}
			Substitutes C APIs used for writing output (e.g.,
			\code{printf}) by C++ APIs (e.g., usage of \code{cout}). \\
			& \emph{C-Style option:} Substitutes C++ APIs used for
			writing output (e.g., usage \code{cout}) by C APIs (e.g.,
			\code{printf}). \\
			\addlinespace[2pt]
			Sync-with-stdio transformer & Enable or remove the
			synchronization of C++ streams and C~streams if possible.
			\\
		\end{tabularx}
	\end{subtable}
	\begin{tabularx}{\textwidth}{p{.21\textwidth}X}
		\bottomrule
	\end{tabularx}
		\vspace{2.7cm}
\end{table*}

\begin{table*}[p]
	\ContinuedFloat
	\centering
	\footnotesize
	\caption{List of Code Transformations (continued)}
	\label{tab:appendix-transf2}
	\begin{subtable}{\linewidth}
		\begin{tabularx}{\textwidth}{p{.21\textwidth}X}
			\toprule
		\end{tabularx}

	\caption*{\bf Template Transformers}
	\begin{tabularx}{\textwidth}{p{.21\textwidth}X}
		\head{Transformer} & \head{Description of Transformations}
		\\
		\midrule
		Identifier transformer & Renames an identifier, i.e., the
		name of a variable or function. If no template is given,
		default values are extracted from the 2016 Code Jam
		Competition set that was used by \citet{CalHarLiuNar+15}
		and that is not part of the training- and test set.
		We test default values such as \code{T}, \code{t}, $\dots$,
		\code{i}. \\
		\addlinespace[2pt]
		Include transformer & Adds includes at the beginning of the
		source file. If no template is given,
		the most common includes from the 2016 Code Jam Competition
		are used as defaults.\\
		\addlinespace[2pt]
		Global declaration transformer & Adds global declarations
		to the source file. Defaults are extracted from the 2016
		Code Jam Competition.\\
		\addlinespace[2pt]
		Include-typedef transformer & Inserts a type using
		\code{typedef},
		and updates all locations where the new type can be
		used. Defaults are extracted from the 2016
		Code Jam Competition.\\
	\end{tabularx}

\end{subtable}
\newline\bigskip\newline
\begin{subtable}{\linewidth}

	\caption*{\bf Miscellaneous Transformers}
	\begin{tabularx}{\textwidth}{p{.21\textwidth}X}
		\head{Transformer} & \head{Description of Transformations}
		\\
		\midrule
		Compound statement transformer & \emph{Insert option:} Adds
		a compound statement (\code{\{...\}}). The transformer adds
		a new compound statement to a control statement (\code{if},
		\code{while}, etc.) given their body is not already wrapped
		in a compound statement.  \\
		& \emph{Delete option:} Deletes a compound statement
		(\code{\{...\}}). The transformer deletes compound
		statements that have no effect, i.e., compound statements
		containing only a single statement. \\
		\addlinespace[2pt]
		Return statement transformer & Adds a return statement. The
		transformer adds a return statement to the main function to
		explicitly return 0 (meaning success). Note that main is a
		non-void function and is required to return an exit code.
		If the
		execution reaches the end of main without encountering a
		return
		statement, zero is returned implicitly. \\
		\addlinespace[2pt]
		Literal transformer & Substitutes a return statement
		returning an integer literal, by a statement that returns a
		variable. The new variable is declared by the transformer
		and initialized accordingly. \\
	\end{tabularx}

	\end{subtable}
	\begin{tabularx}{\textwidth}{p{.21\textwidth}X}
		\bottomrule
	\end{tabularx}
	\vspace{12.2cm}
\end{table*}

\end{document}